# Unsupervised Fault Detection using SAM with a Moving Window Approach

Ahmed Maged [a,] Herman Shen [a*]
[a] *Department of Mechanical Engineering, University of North Texas, Denton, Texas, USA, 76207*
*\* Corresponding Author;* Herman.Shen@unt.edu

**Abstract**

Automated fault detection and monitoring in engineering are critical but frequently difficult owing to the necessity for collecting and labeling large amounts of defective samples. We present an unsupervised method that uses the high-end Segment Anything Model (SAM) and a moving window approach. SAM has gained recognition in AI image segmentation communities for its accuracy and versatility. However, its performance can be inconsistent when dealing with certain unexpected shapes, such as shadows and subtle surface irregularities. This limitation raises concerns about its applicability for fault detection in real-world scenarios. We aim to overcome these challenges without requiring fine-tuning or labeled data. Our technique divides pictures into smaller windows, which are subsequently processed using SAM. This increases the accuracy of fault identification by focusing on localized details. We compute the sizes of the segmented sections and then use a clustering technique to discover consistent fault areas while filtering out noise. To further improve the method's robustness, we propose adding the Exponentially Weighted Moving Average (EWMA) technique for continuous monitoring in industrial settings, which would improve the method's capacity to trace faults over time. We compare our method to various well-established methods using a real case study where our model achieves 0.96 accuracy compared to 0.85 for the second-best method. We also compare our method using two open-source datasets where our model attains a consistent 0.86 accuracy across the datasets compared to 0.53 and 0.54 for second-best models.

## 1. Introduction

It is inevitable to encounter surface defects in many industrial applications, including but not limited to semiconductors [1], fabrics [2], and steel [3]. These defects have the potential to change the material properties and cause industrial accidents on top of bad user experiences [4]. For example, flaws in the surface of steel can significantly reduce its contact fatigue strength, leading to premature failure of components. Hence, surface quality inspection is one of the important aspects of industrial production.

Over the past decades, many surface defect detection methods have been proposed, which can be divided into two categories: traditional methods and deep learning-based methods. Conventional methods rely on manual feature extraction (e.g., texture-based, color-based, and shape-based features) and threshold setting to detect defects. These methods usually have limited feature extraction capabilities and poor robustness. In contrast, deep learning methods, which are data-driven, can automatically extract relevant features by training on large datasets. They have strong feature extraction capabilities and show good generalization [5].

The majority of deep learning-based methods for surface defect detection are supervised learning approaches[6] [7], [8], [9]. These methods show good performance, yet they require a significant number of defective/non-defective samples and their corresponding labels, which can be challenging to obtain in industrial settings. In real-world settings, fault-free samples outnumber defective ones. A worse situation could emerge if newer defect types appear throughout the production process. Further, labeling such large data necessitates the expertise of trained engineers or technicians, which takes considerable time and effort. For such reasons, the practicality of supervised learning in the industrial sector is quite limited [10].

Unsupervised learning methods have become increasingly popular since they only train on unlabeled data [11]. During the training process, they learn the underlying distribution from these normal samples and accordingly identify samples that are significantly different from this learned distribution as potential anomalies. Autoencoders (AEs) [12] and Generative Adversarial Nets (GANs) [13] along with their variants have been used to train defect detection models only on defect-free training samples. Popular variants of the AEs used in defect detection tasks include the variational autoencoder (VAE) [14] and the convolutional variation autoencoder (CVAE) [15], among others. AE and its variants-based methods typically feed the latent features directly to the decoder, leading to a representation of the latent space that is sometimes under-designed**.** Hence, anomaly detection can be achieved by further processing the latent space. Examples of methods based on AE include AE-SSIM, a defect inspection method using structural similarity with an AE [16]. Another similar method is Deep Embedded Clustering (DEC), developed by [17]. It simultaneously learns feature representations and cluster assignments using deep neural networks. DEC learns a mapping from the data space to a lower-dimensional feature space in which it iteratively optimizes a clustering objective. Likewise, Zhang et al. [18] proposed Vector

Quantized-Variational Autoencoder (VQ-VAE2) for unsupervised anomaly detection for stamped metal products. The method is able to reconstruct input images while retaining crack details. The latent features at different scales are quantized into discrete representations then they use an autoregressive model to learn the distribution of these discrete representations. A different approach is control charts-based monitoring using VAE [19]. VAE-based control chart uses VAE to find a latent space that is normally distributed, and then that space is monitored using $T^2$ and $SPE$ chart. Similarly, Maged et al. [20] proposed a VAE-LSTM chart where the latent space is fed into another Long Short-Term Memory (LSTM) network, and the residuals are monitored later using $T^2$ chart. Pen et al. [21] modified the VAE monitoring by proposing an Interpretable Latent Variable Model. Prifti et al. [17] suggested using a Convolutional Neural Network-VAE (CVAE) as an unsupervised method for anomaly detection in Scanning Transmission Electron Microscopy (STEM). Similarly, Yun et al. [23] proposed Conditional CVAE for fault detection for metal surfaces in situations where data imbalance is encountered.

On the other hand, GANs can formulate the latent space using adversarial learning of the representation of samples, whereas the goal of GAN is to achieve a balance between the generator and the discriminator, then the latent space is further processed by different means. For example, Schlege et al. [24] introduced AnoGAN, an anomaly detection generative adversarial network that learns the distribution of defect-free texture image patches using GAN techniques. It detects defects by finding a latent sample that reproduces a given input image patch. Anomaly [25] employs a conditional GAN for anomaly detection through an encoder-decoder–encoder generator framework. Skip-GANomaly [26] enhanced this approach by adding a skip connection to improve the reconstruction quality of image backgrounds. Although these GAN-based methods are effective at detecting and localizing various defects, they face challenges in balancing noise-free normal background reconstruction with precise defect localization.

An alternative approach is to use pretrained models combined with unsupervised classification methods. Heckler et al [27] explored the importance of pretrained feature extractors for unsupervised anomaly detection. In this regard, one should use a pretrained network such as ResNet or EfficientNet to extract relevant features, then apply an unsupervised classification method such as K-means or DBSCAN to distinguish the anomalous instances from the non-anomalous ones effectively.

To address these challenges, we present an unsupervised fault detection and monitoring model of surface images based on the Segment Anything Model (SAM) [28]. SAM is a state-of-the-art segmentation tool chosen for its remarkable segmentation accuracy and robustness due to training on a significant volume of data, thereby removing any necessity for fine-tuning. The images are divided into sub-images of small size or windows with a moving step, which are then processed through SAM one at a time. This ensures better segmentation results since SAM performs better with clearer images. The moving windows are user-defined based on the anticipated defect size. This individualization enables one to have control with high exactness over the process of detection. There are four main advantages to our proposed method:

- Using SAM with a moving window approach enhances segmentation precision by focusing on smaller sub-images, ensuring better correlation between pixels within each original image region.
- No training is required on the collected data since SAM is known for its exceptional accuracy.
- We propose a clustering algorithm with tolerance that improves defect detection by filtering out noise and identifying consistent defect regions.
- The model is highly explainable, meaning we can focus on localized details since we can easily examine/review the questionable segmented image with the masks overlaid on the original image.

To further enhance the method's applicability in industrial settings, we propose integrating the Exponentially Weighted Moving Average (EWMA) scheme to track defect trends over time. This can enhance the proposed method's applicability in industrial settings where we are interested in detecting minor shifts as well as lowering the number of false alarms.

The rest of the article is organized as follows, In Section 2 we introduce our new method, including the SAM-based fault detection with a moving window approach and clustering algorithm. Then in Section 3, we present our experimental results and comparative studies. Later, in Section 4, we discuss the results and address the limitations of the proposed method. We also discuss the reasoning for parameter selection in the same section. Finally, Section 5 concludes the article and provide future suggestions.

## 2. Proposed Methodology

In the context of engineering and machine vision, detecting and monitoring faults on surfaces is crucial. Our model addresses this by utilizing an unsupervised method for fault detection and monitoring of surface images. As we mentioned earlier, the core of our approach is the SAM, selected for its exceptional segmentation accuracy and robustness. It is derived from being trained on a large dataset, which eliminates the need for fine-tuning. A presentation of the general architecture of SAM is presented in Figure 1.

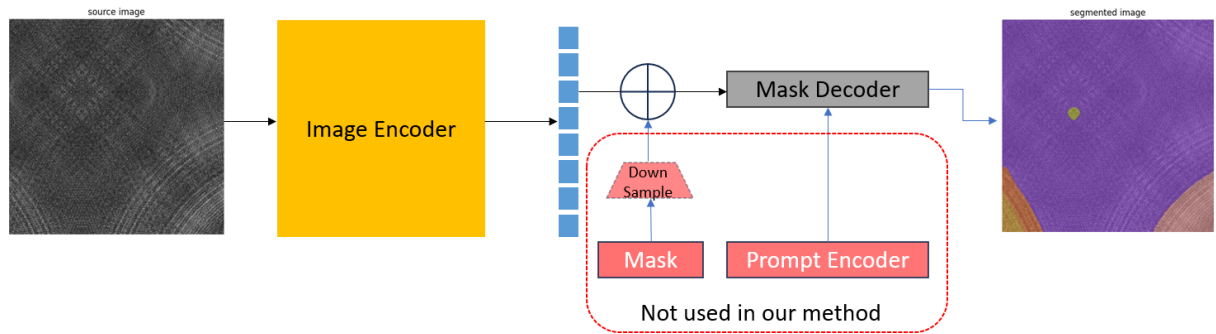

Figure 1. Architecture of Segment Anything Model (SAM)

As shown in Figure 1, SAM consists of three components: an image encoder, a prompt encoder, and a mask decoder [28]. The image encoder part is a pretrained Vision Transformer (ViT) called Masked AE [29]. The prompt encoder can be in the form of one of four methods: points, boxes, text, or masks. The mask decoder efficiently maps the image embeddings, and prompt embeddings and outputs a token to a mask. Although we do not use any prompts in this paper, we believe that including prompts in future research could significantly enhance defect detection (further discussion on this is provided in the Conclusion section). SAM is trained on the Segment Anything 1 Billion Mask (SA-1B) dataset, which contains over 1.1 billion high-quality segmentation masks derived from 11 million images. This is arguably the largest dataset for image segmentation tasks to date. Moreover, this dataset was carefully curated to cover a wide range of domains, objects, and scenarios, including medical imagery, satellite images, and more [28]. The diversity of the SA-1B dataset helps the model generalize well across different tasks. The sheer size and variety of the dataset provided ample training data for the model to learn complex patterns and representations. Accordingly, this enables SAM to achieve state-of-the-art performance on diverse segmentation tasks, often surpasses previous fully supervised results in a zero-shot manner (i.e., without task-specific fine-tuning).

So, SAM can be very efficient in segmenting defects on any image where defect masks can be utilized as a source for identifying anomalies on surfaces. However, solely relying on SAM can produce numerous false results. This is because SAM sometimes generates unwanted segmentation masks (See Figure 2). This could be adjusted through fine-tuning, but this is only occasionally feasible due to the need for labeled data. Accordingly, we enhance the performance of SAM for fault detection through the proposed approach as proposed in Figure 3.

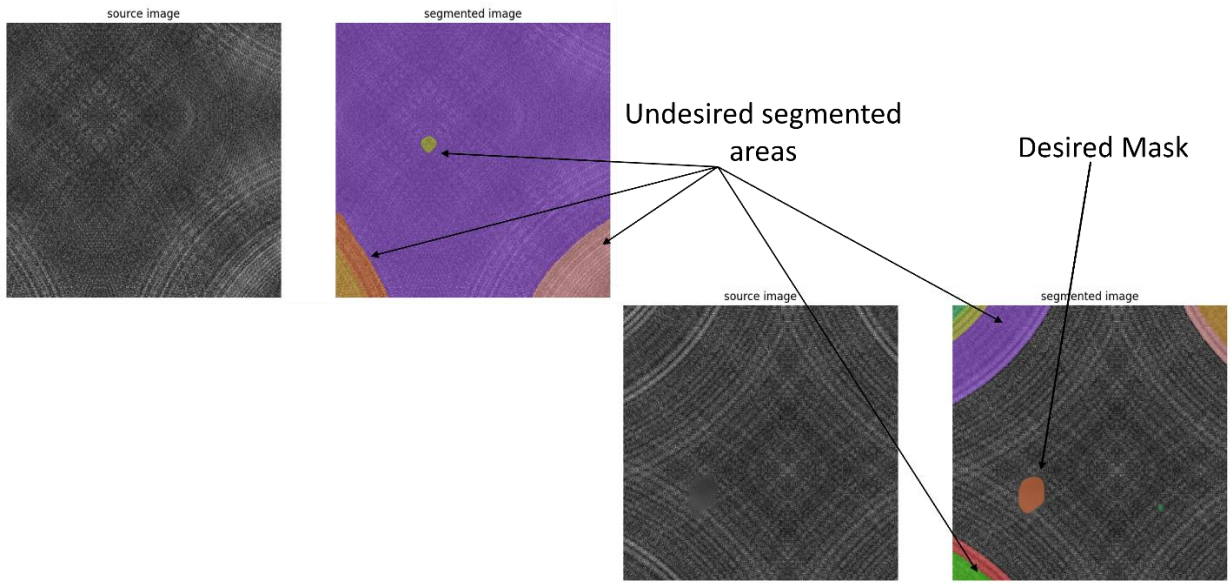

Figure 2. Undesired Segmented masks when processing images

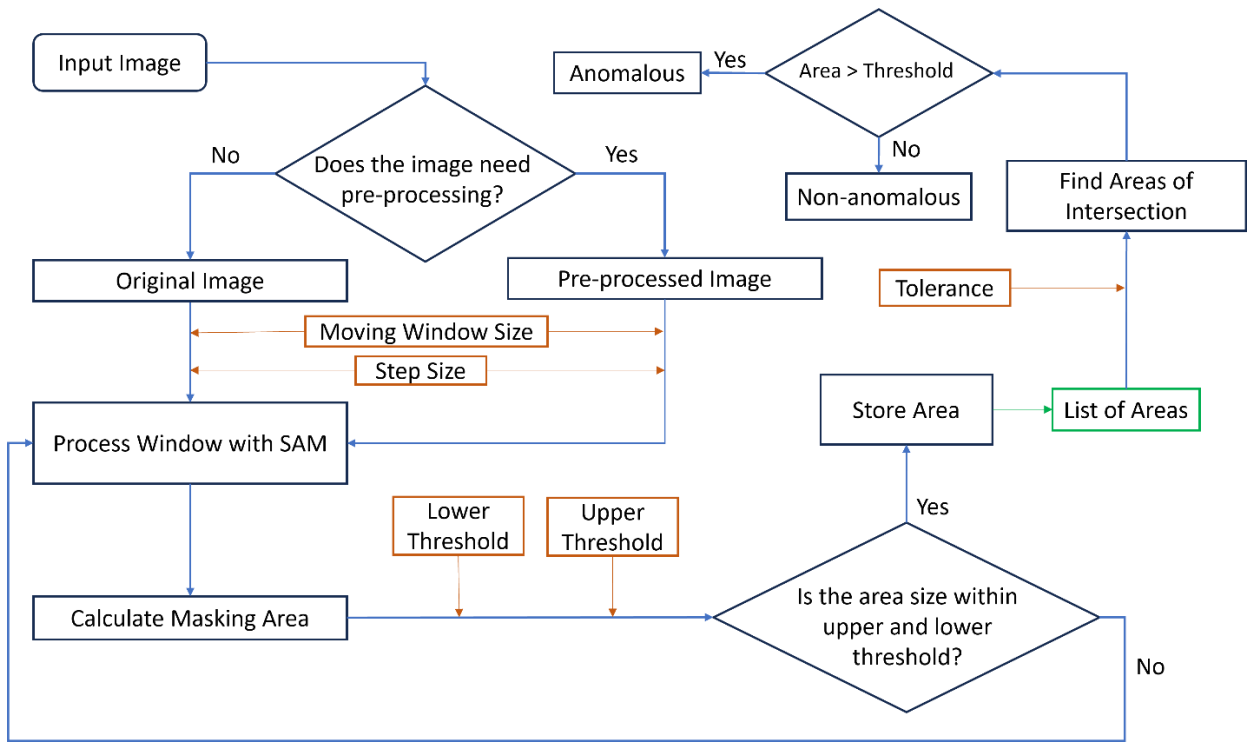

Figure 3. Flow chart for the proposed method

First, the image is checked for preprocessing needs (e.g., hue, saturation, brightness), then we implement a moving window approach as illustrated in Algorithm 1. Each image is divided into smaller sub-images, or windows, which are processed individually through SAM one at a time. This approach ensures better segmentation results since SAM performs optimally on smaller images. For illustration, we show actual segmentation results for a full image and a sub-image from it in Figure 4. It is evident that outcomes change when processing the entire image versus smaller windows. Note that the window size and the step size for moving the windows are user-defined, ensuring defects appear in multiple windows. This individualization enables one to control the detection process with high exactness.

| **Algorithm 1:** | Split Image. |
|---|---|
| **Input** | $Image, Window\ width, Window\ height, Width\ step, Height\ step$ |
| **Output** | $Windows$ |
| | 1:  $height, width \leftarrow \text{Dimensions}(Image)$ |
| | 2:  $Windows \leftarrow \emptyset$ |
| | 3:  **for** $i \leftarrow 1$ **to** $height - Window\ height + 1$ **step** $Height\ step$ **do** |
| | 4:   **for** $j \leftarrow 1$ **to** $width - Window\ width + 1$ **step** $Width\ step$ **do** |
| | 5:    $window \leftarrow image[i : i + Window\ height, j : j + Window\ width]$ |
| | 6:    $Windows \leftarrow Windows \cup \{window\}$ |
| | 7:   **end for** |
| | 8:  **end for** |

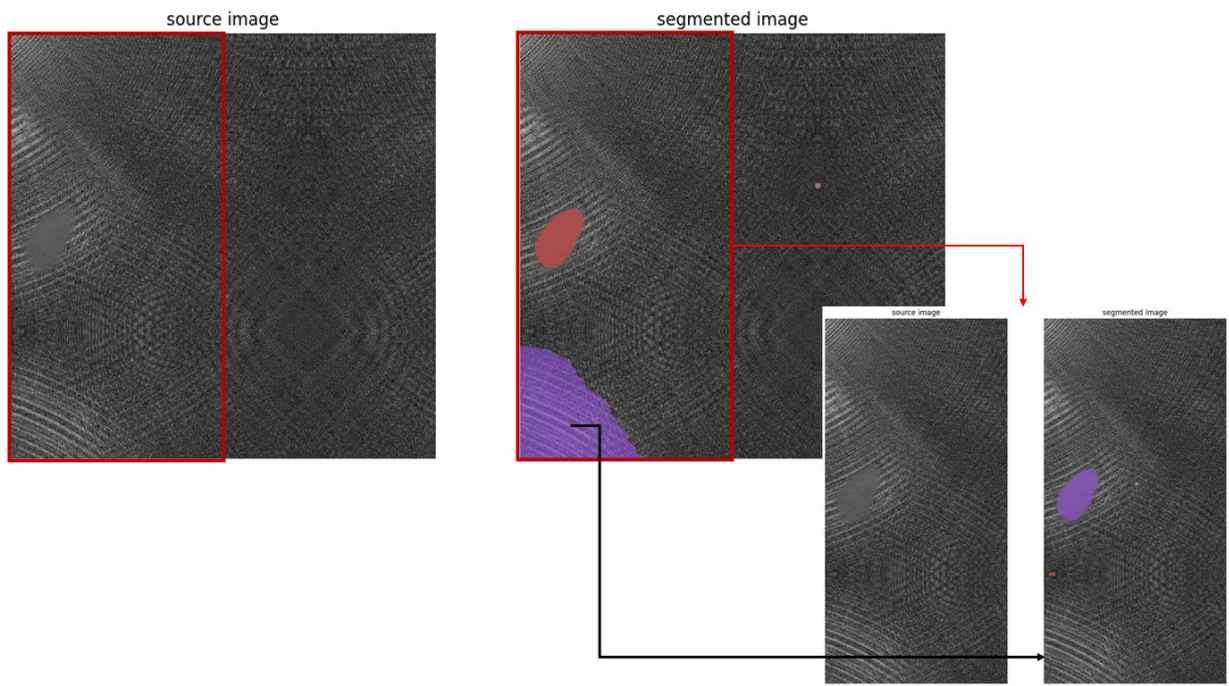

Figure 4. Undesired mask disappeared after processing a smaller window

SAM processes each sub-image and outputs segmentation masks as implemented in Algorithm 2. We calculate the areas of these segmented regions to quantify the significant areas that might indicate defects. The following step is crucial as it helps differentiate between defects and noise. Noise masks occur since we process smaller windows, which can sometimes capture irrelevant details or artifacts, as illustrated in Figure 5. To address this issue, we introduce a thresholding mechanism presented in Algorithm 3. Masks with areas falling outside predefined upper and lower thresholds are filtered out, ensuring only significant regions are considered. This filtering helps reduce false positives and focuses the analysis on potential defective areas.

| Algorithm 2: | Window Segmentation. |
|---|---|
| **Input** | $Windows$, SAM model |
| **Output** | $Masks$ |
| | 1:    $Masks \leftarrow \emptyset$ |
| | 2:    **for each** $window$ **in** $Windows$ **do** |
| | 3:       $sam\ result \leftarrow$ **generate masks using SAM model** ($window$) |
| | 4:       $window\_mask \leftarrow$ **extract** $mask$ **from** $sam\ result$ |
| | 5:       $Masks \leftarrow Masks \cup \{window\_mask\}$ |
| | 6:    **end for** |

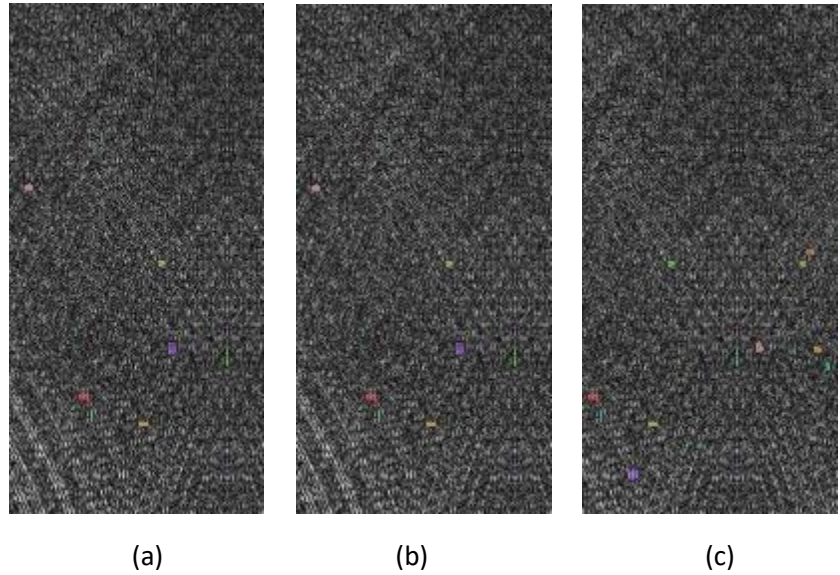

        (a)                      (b)                      (c)

Figure 5. Exemplary undesired masked noise appearing at three different windows a, b and c

| Algorithm 3: | Calculate Masks' Areas |
|---|---|
| **Input** | $Masks$, $Upper\ Threshold$, $Lower\ Threshold$ |
| **Output** | $Masks'\ Areas$ |
| | 1:    $Masks\ Areas \leftarrow \emptyset$ |
| | 2:    **for each** $mask$ **in** $Masks$ **do** |
| | 3:       $mask\ width, mask\ height = mask[1], mask[2]$ |
| | 4:       $mask\ area\ =\ mask\ width \times mask\ height$ |
| | 5:       **if** $Lower\ Threshold\ <\ mask\ area\ <\ Upper\ Threshold$ **then** |
| | 6:          $Masks\ Areas \leftarrow Masks\ Areas \cup \{mask\ area\}$ |
| | 7:       **end if** |
| | 8:    **end for** |

Then, we employ an adaptive clustering algorithm to further refine the detection process. This algorithm groups similar segmented areas based on a defined tolerance level, identifying clusters that likely represent actual defects. By calculating the intersection of these clusters, we pinpoint consistent defect regions across multiple windows, enhancing the reliability of our detection method. Algorithm 4 identifies the most frequent area, representing the defect, by clustering the detected areas and finding the largest cluster, ensuring accurate pinpointing of the defective area. Calculating the area of intersection is a critical part of our method since it will be checked against a threshold. Note that we hypothesize that defects will be detected consistently across different windows; we can differentiate between actual defects and noise. Figures 6 and 7 show holistically how our proposed method works.

In these figures, the source image is divided into smaller sub-images or windows and then processed individually by SAM to produce segmented images. In Figure 6, the segmented image on the right demonstrates how processing the full image produces undesired masks. Then, we employ our approach, using different windows to capture various regions of interest. Some windows do not show significant intersections, which indicates that this is possibly a non-defective region. Similarly, other regions either exceed or fall below the upper and lower thresholds, respectively indicating potential undesired segmentation masks. This visualization highlights the effectiveness of the moving window approach in isolating and identifying defects that might be missed when analyzing the whole image at once. Note that the ground truth of this figure is Fault Free.

Similarly, Figure 7 shows another example; nevertheless, this one represents a faulty image. We can see that when the windows are processed through SAM, the potential defective area remains through them revealing clear intersections. While other undesired masks have higher-than-threshold areas in others, and small intersections in yet another set. This differentiation is crucial for accurately locating defect regions and reducing noise.

| **Algorithm 4:** | Find Intersections |
|---|---|
| **Input** | $Masked\ areas, tolerance$ |
| **Output** | $Intersection$ |
| | 1:    $sorted\ areas \leftarrow sort(Masked\ areas)$ |
| | 2:    $area\ clusters \leftarrow \emptyset$ |
| | 3:    $current\ area\ cluster \leftarrow \{sorted\ areas[1]\}$ |
| | 4:    **for each** $area$ **in** $sorted\ areas$ **do** |
| | 5:      **if** $abs(area - mean(current\ area\ cluster)) \leq tolerance$ **then** |
| | 6:        $current\ area\ cluster \leftarrow current\ area\ cluster \cup \{area\}$ |
| | 7:      **else** |
| | 8:        $area\ clusters \leftarrow area\ clusters \cup \{current\ area\ cluster\}$ |
| | 9:        $current\ area\ cluster \leftarrow \{area\}$ |

```
10:        end if
11:    end for
12:    area clusters ← area clusters ∪ {current area cluster}
13:    Intersection ← arg max sum(area clusters)
14:    if length(Intersection) > 1 then
15:        Intersection = mean(Intersection)
16:    else
17:        Intersection = 0
18:    end if
```

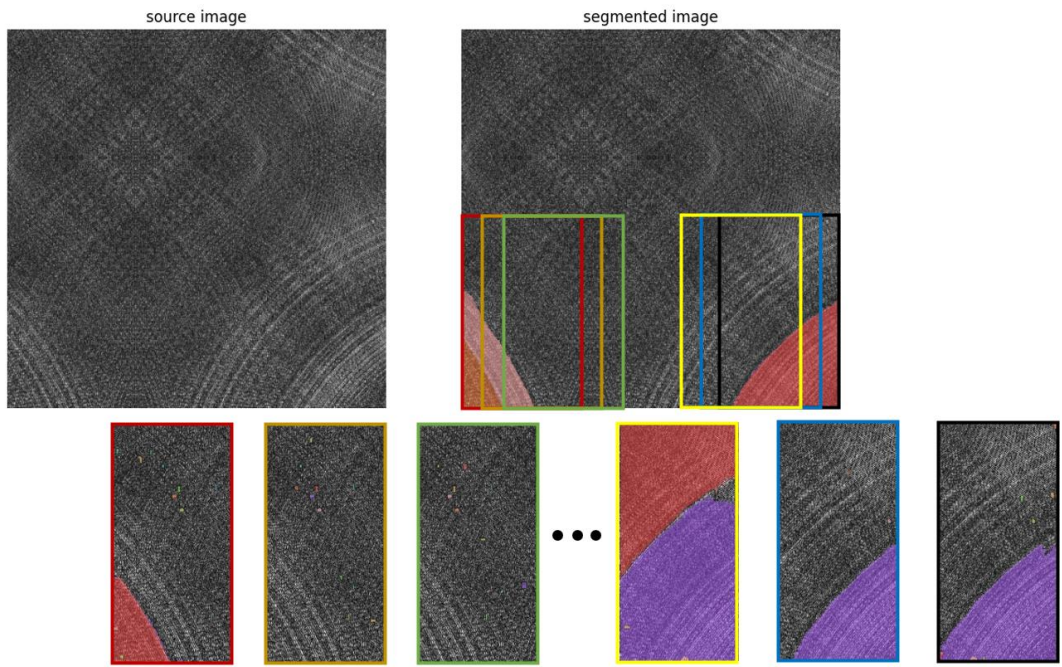

Figure 6. Proposed segmentation method results applied to a fault-free sample

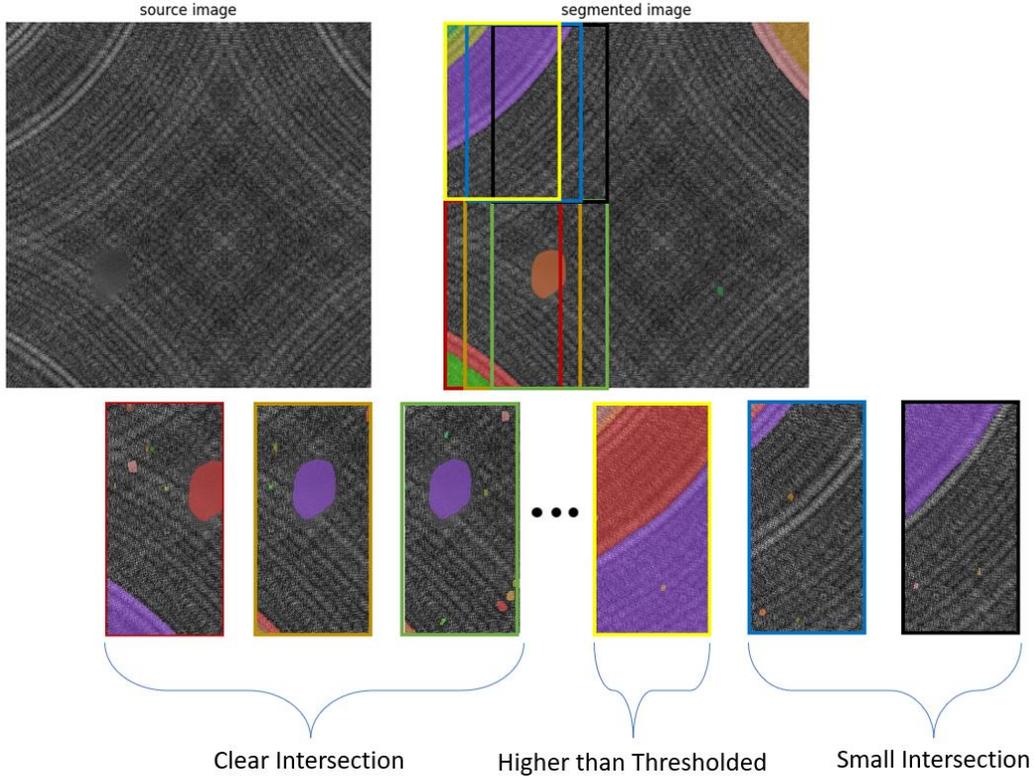

Figure 7. Proposed segmentation method results applied to a faulty sample

One further step that can significantly enhance our method's applicability in industrial settings is using the Exponentially Weighted Moving Average (EWMA) scheme. EWMA is an effective statistical technique that continuously monitors processes for changes. Our proposed EWMA monitoring statistic $Z_t$ with smoothing parameter $\lambda$ is given as

$$Z_t = (1 - \lambda)Z_{(t-1)} + \lambda x_t$$

where $z_0 := E_{fault\ free}(X)$, $0 < \lambda < 1$, and $x_t$ is the current observation (masked area). By giving higher weights to the more recent observations, EWMA is sensitive to recent trends and changes in the data compared to a simple use of the individual values. This sensitivity is crucial for early detection of emerging defects in industrial applications. Note that EWMA monitoring is more effective under the assumption that process shifts happen gradually over time, which is the common case in industry. In this paper, we suggest using EWMA so that we can detect minor defect pattern changes, provide a robust and reliable solution, as well as lower the false alarm rates. The continuous feedback loop from EWMA enhances our model's fault detection capabilities. This integration offers a scalable and efficient solution for industries that require constant monitoring and high precision in fault detection. The threshold, also called Upper Control Limit (UCL) is calculated based on the 95% empirical quantiles for the in-control variables based on Empirical Cumulative Distribution Function (ECDF).

## 3. Case Studies and Performance Evaluation

In this section, we apply the proposed method to a real-world dataset from industry and two open-source datasets. We compare the performance of our method against several well-known unsupervised anomaly detection methods, including CVAE-T² Chart, CVAE-SPE Chart, Resnet-Kmeans, Resnet-DBSCAN, AE-SSIM, DEC, AnoGAN, and GAN-kmeans. Performance is evaluated using five metrics: Accuracy, Precision, Recall, F1-Score, and AUROC. Accuracy is calculated as $\frac{TP+TN}{TP+TN+FP+FN}$, Precision as $\frac{TP}{TP+FP}$, Recall as $\frac{TP}{TP+FN}$, F1-Score as $2 \times \frac{Precision \times Recall}{Precision + Recall}$, and AUROC is the measured area under the Receiver Operating Characteristic (ROC) curve. We also provide the confusion matrix for a detailed performance analysis.

### 3.1 Case Study

In this case study, we examine a textile manufacturing process. The producer manufactures high-quality textiles, ensuring strict quality control measures. We focus on detecting surface faults in the textile production process using fabric images. In brief, the manufacturing process begins with the preparation of raw materials, where natural or synthetic fibers are cleaned and spun into yarn. The yarn is then woven into the fabric using advanced weaving machines, determining the texture and strength of the fabric. Following this, the woven fabric undergoes dyeing and finishing processes to achieve the desired colors and properties such as softness, water resistance, or durability.

Quality control is a critical stage in this process, where the fabrics are inspected for defects. To maintain high standards, the factory employs image analysis for fault detection. The process involves capturing images of the fabric at different stages of production and analyzing them for any defects. Examples of fault-free and faulty samples are shown in Figures 8 and 9. It can be noticed that the anomalies in the faulty images are very subtle, which means it can be very hard to be detected by usual methods. The dataset contains 100 images, 50 of them represent fault-free images and 50 others represent the faulty class.

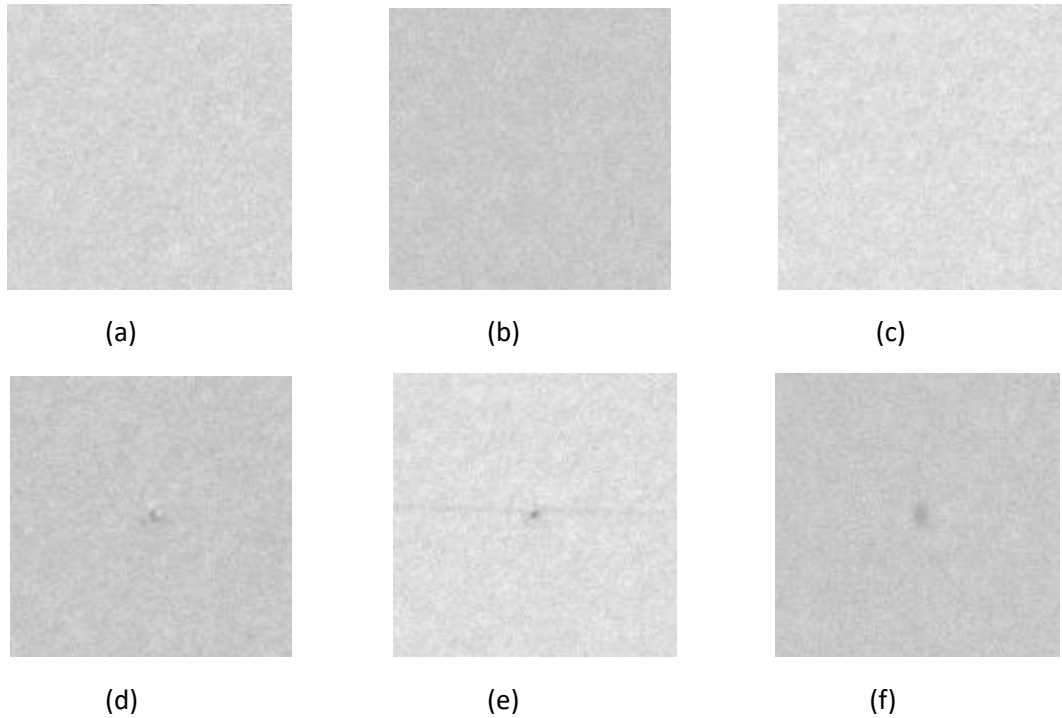

(a) (b) (c)

(d) (e) (f)

Figure 8. Sample images from the case study, images a, b, c represent fault free samples, and images d, e, f represent faulty samples

We apply the proposed method to the dataset and report the results in Table 1. Based on the results from the table, several observations can be made about the performance of each method. The proposed method demonstrates the highest overall performance, with an accuracy of 0.96, perfect precision of 1.00, recall of 0.92, an F1-score of 0.96, and an AUROC of 0.960784. This indicates an excellent balance of precision, recall, and discriminative power. Resnet-DBSCAN also performs well, achieving an accuracy of 0.85, perfect precision of 1.00, recall of 0.71, and an F1-score of 0.83, with an AUROC of 0.85. However, it is outperformed by the proposed method in both recall and overall accuracy. Resnet-Kmeans has a moderate accuracy of 0.6765 and significantly lower recall of 0.3529, resulting in a lower F1-score of 0.5217 despite its perfect precision. AE-SSIM shows good accuracy at 0.69, and perfect precision, but its recall is lower at 0.37, leading to an F1-score of 0.54 and an AUROC of 0.69.

The CVAE-based methods, CVAE-T² Chart, and CVAE-SPE Chart, both exhibit low accuracies of 0.59 and 0.57 respectively. They have high precision (0.91 and 0.82) but very low recall (0.20 and 0.18), resulting in low F1-scores of 0.32 and 0.29, and moderate AUROC values of 0.70 and 0.74. AnoGAN performs poorly across the board, with the lowest accuracy of 0.4902, a very low recall of 0.04, and an F1-score of 0.0714, alongside the lowest AUROC of 0.291811. Similarly, GAN-kmeans also shows low performance with an accuracy of 0.5, precision and recall of 0.5, and a low F1-score of 0.3792, with an AUROC of 0.5, indicating performance no better than random chance. The superiority of the proposed model is also evident in the confusion matrix in Figure. 9.

Table 1. Comparison across all the used models for the case study

| Method | Accuracy | Precision | Recall | F1-Score | AUROC |
| --- | --- | --- | --- | --- | --- |
| CVAE-$T^2$ Chart | 0.59 | 0.91 | 0.2 | 0.32 | 0.70 |
| CVAE-SPE Chart | 0.57 | 0.82 | 0.18 | 0.29 | 0.74 |
| Resnet-Kmeans | 0.6765 | 1 | 0.3529 | 0.5217 | 0.676471 |
| Resnet-DBSCAN | 0.85 | 1 | 0.71 | 0.83 | 0.85 |
| AE-SSIM | 0.69 | 1 | 0.37 | 0.54 | 0.69 |
| DEC | 0.5098 | 0.5052 | 0.9608 | 0.6622 | 0.509804 |
| AnoGAN | 0.4902 | 0.4 | 0.04 | 0.07 | 0.291811 |
| GAN-kmeans | 0.5 | 0.5 | 0.5 | 0.3792 | 0.5 |
| Proposed Method | **0.96** | 1 | **0.92** | **0.96** | **0.960784** |

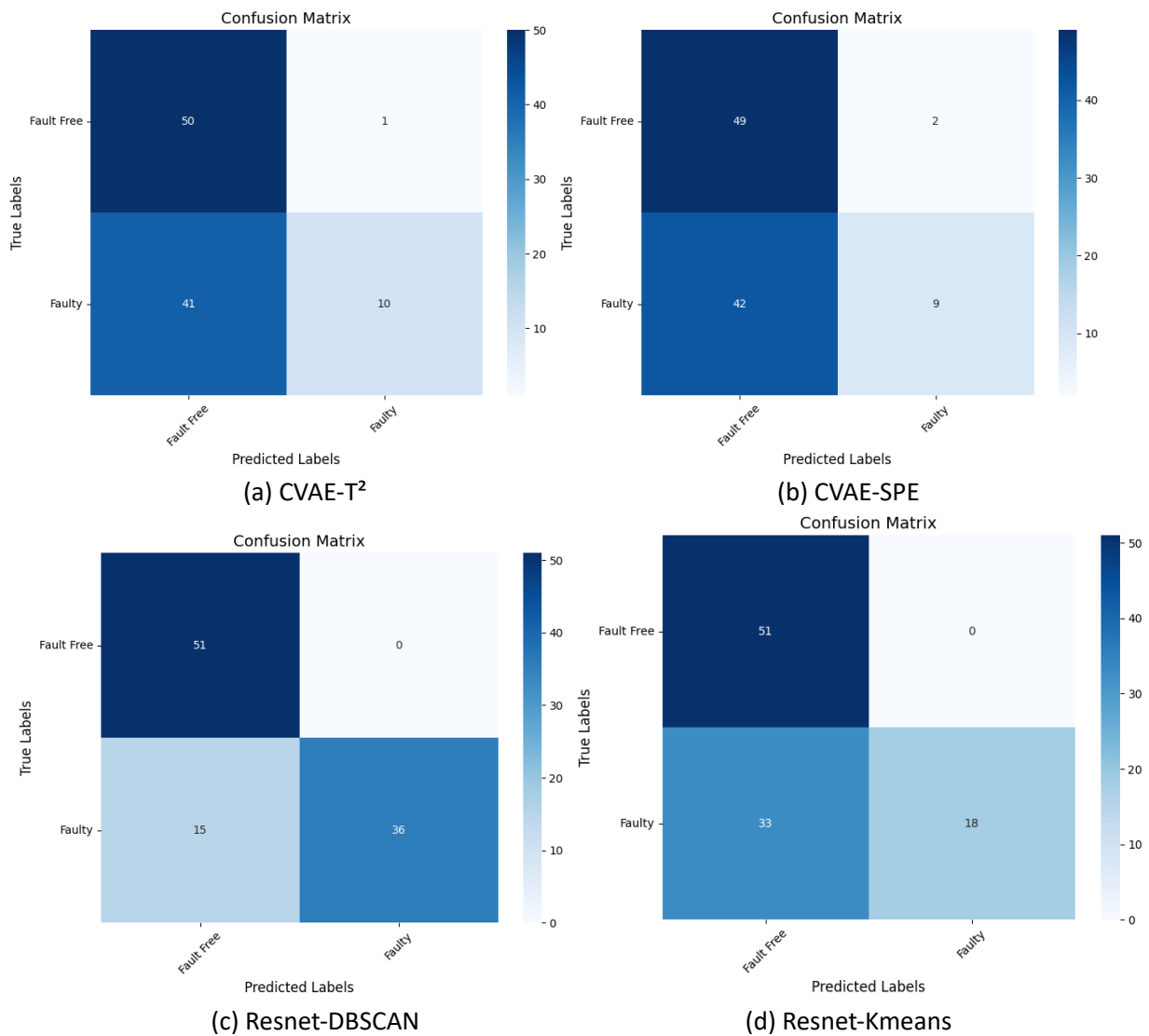

(a) CVAE-$T^2$  (b) CVAE-SPE

(c) Resnet-DBSCAN  (d) Resnet-Kmeans

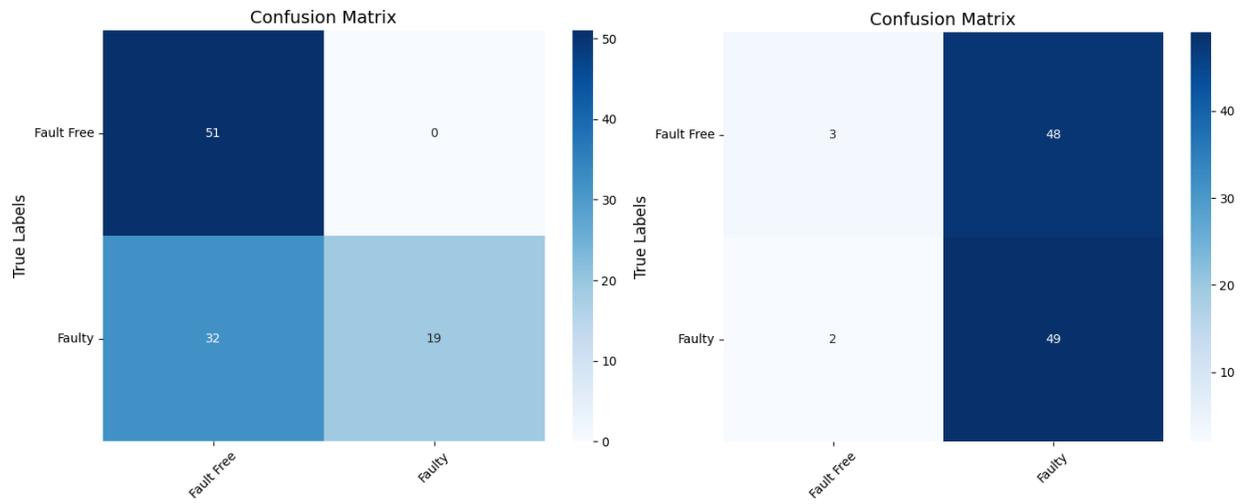

(e) AE-SSIM　　　　　　　　　　　　　(f) DEC

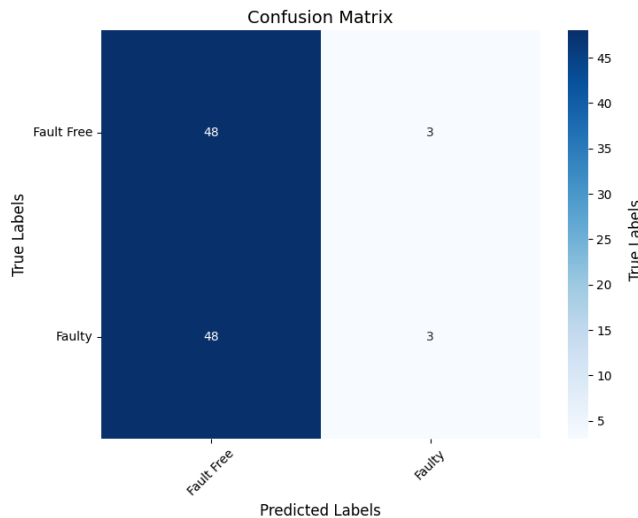

(g) GAN-kmeans　　　　　　　　　　　(h) AnoGAN

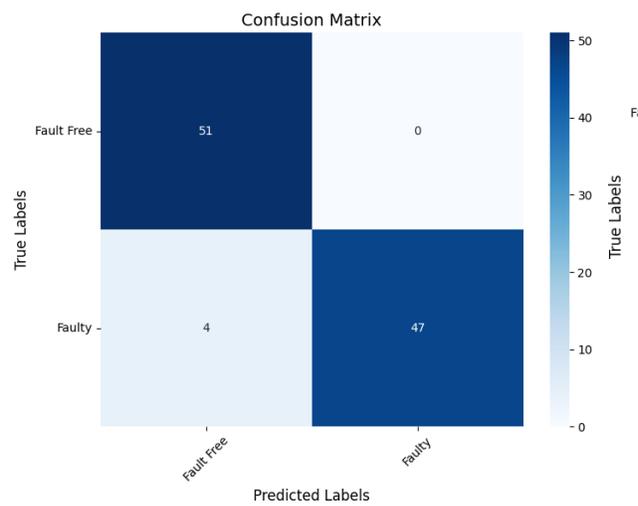

(I) Proposed method　　　　　　　　　(j) Proposed method + EWMA

Figure 9. Confusion Matrix for all the models applied to the case study

In case we assume a monitoring strategy and apply the EWMA scheme, the performance metrics improve to perfect scores across the board: 1.00 for accuracy, precision, recall, F1-score, and AUROC as shown in Table 2. This indicates that the addition of EWMA significantly enhances the model's ability to correctly classify all instances without any errors, achieving perfect classification performance. However, this relies on the assumption of real-time monitoring of a real process where we inherently assume common cause and special cause variations.

Table 2. Effect of using EWMA on the performance of the proposed model for the case study

| Method | Accuracy | Precision | Recall | F1-Score | AUROC |
|---|---|---|---|---|---|
| Proposed Method | 0.96 | 1 | 0.92 | 0.96 | 0.960784 |
| Proposed method + EWMA | 1 | 1 | 1 | 1 | 1 |

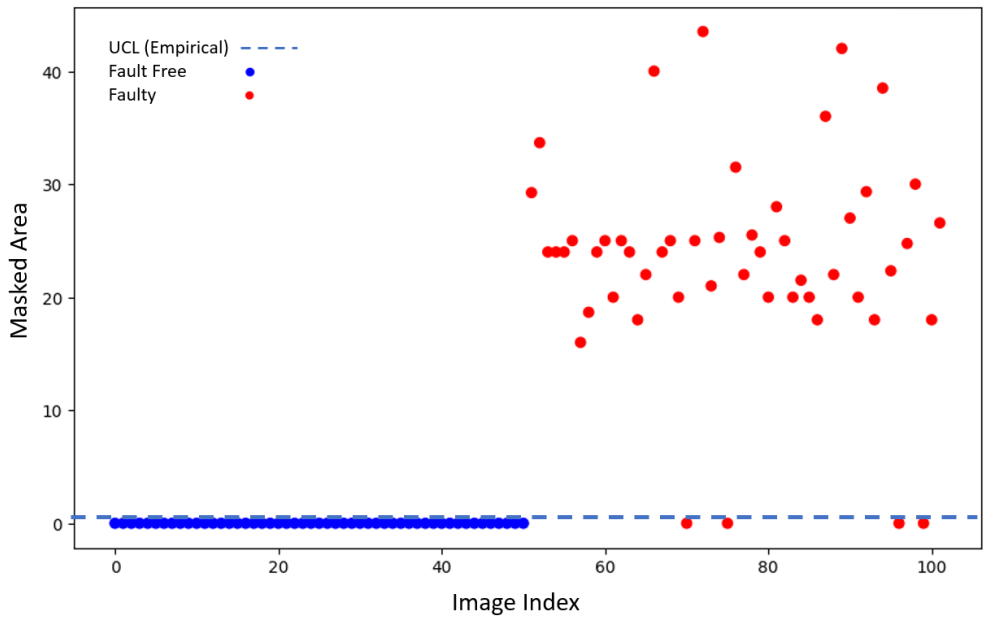

(a) Proposed method

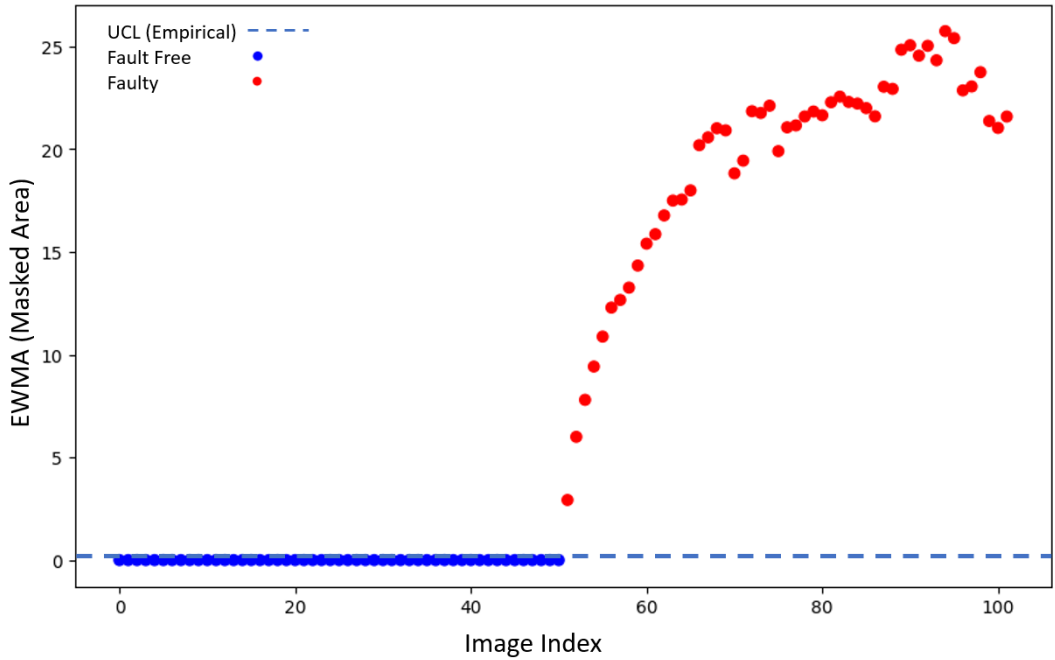

(b) Proposed method + EWMA

Figure 10. Monitoring using the proposed method and EWMA

### 3.2 Open-Source Dataset I

This dataset comes from challenges in industrial image processing, where it is usually used as a benchmark corpus for defect detection on statistically textured surfaces using of industrial optical inspection [1]. While the dataset is typically used with provided labels for supervised learning, we use it for unsupervised learning tasks, where the algorithm learns to identify defects without relying on those labels. The dataset consists of various sub-datasets, and we used two of them to demonstrate the efficiency of our proposed method. A sample image from the dataset is shown in Figure 11. For a fair comparison, we use 150 images from each class. Although our model is unaffected by the balancing problem, we aim to demonstrate its efficiency against other models which might inherently require the data to be balanced.

---

[1] Dataset can be found in https://hci.iwr.uni-heidelberg.de/content/weakly-supervised-learning-industrial-optical-inspection

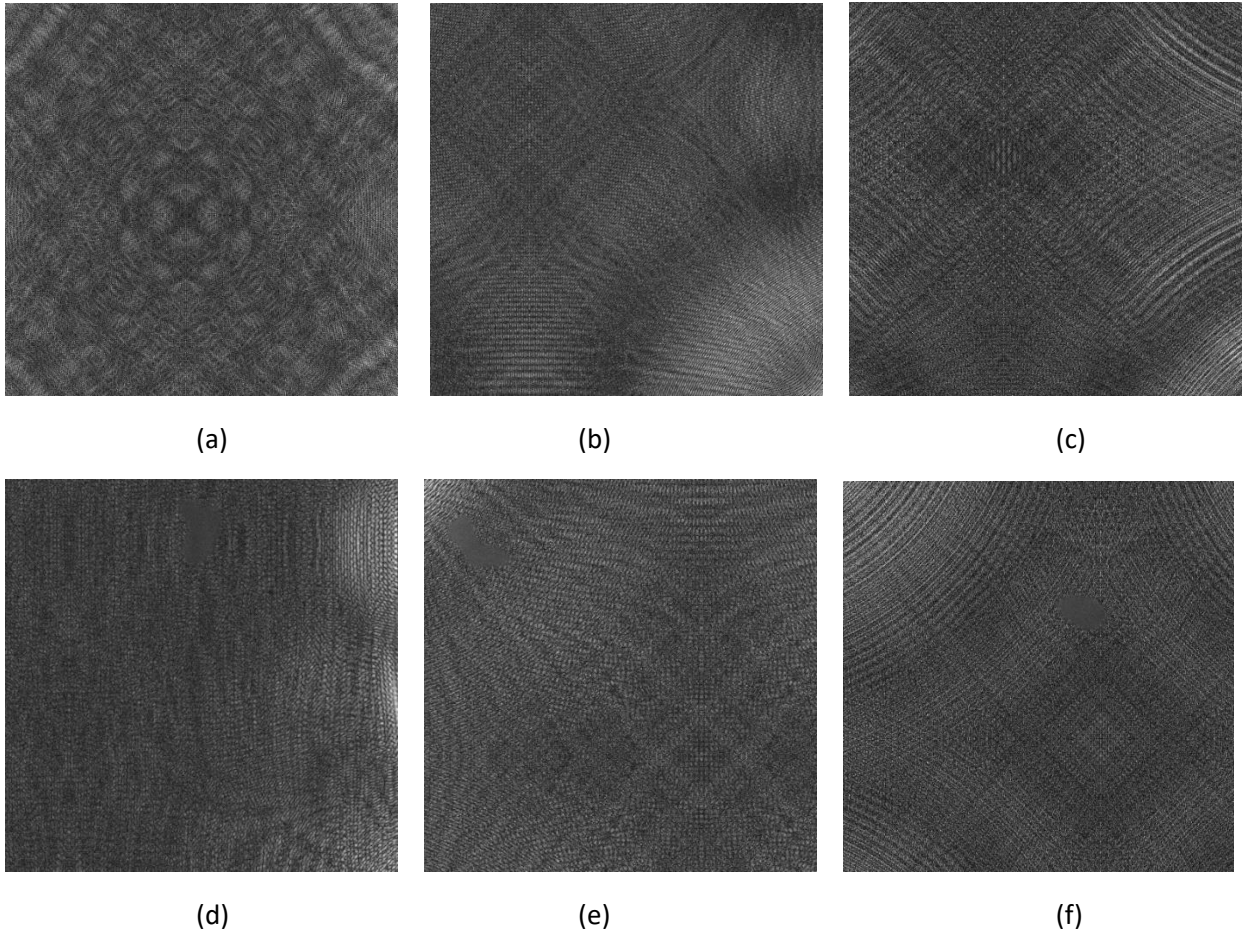

Figure 11. Sample images from dataset I, images a, b, c represent fault free samples, and images d, e, f represent faulty samples

The results shown in Table 3 illustrate that the proposed method significantly outperforms all other methods in all metrics. It achieves an accuracy of 0.86, precision of 0.84, recall of 0.89, F1-score of 0.86, and AUROC of 0.86. Other methods have much lower scores, with the highest precision being 1.0 (Resnet-DBSCAN) but with a low recall of 0.07. The proposed method is the most effective, demonstrating balanced and high performance across all metrics. We can also reach the same conclusion by examining the confusing matrices in Figure 12.

Table 3. Comparison across all the used models for dataset I

| Method | Accuracy | Precision | Recall | F1-Score | AUROC |
|---|---|---|---|---|---|
| CVAE-$T^2$ Chart | 0.52 | 0.58 | 0.12 | 0.2 | 0.51 |
| CVAE-SPE Chart | 0.53 | 0.6 | 0.16 | 0.25 | 0.53 |
| Resnet-Kmeans | 0.51 | 0.51 | 0.56 | 0.53 | 0.51 |
| Resnet-DBSCAN | 0.53 | 1 | 0.07 | 0.13 | 0.53 |
| AE-SSIM | 0.5 | 0.52 | 0.11 | 0.19 | 0.5 |
| DEC | 0.51 | 0.51 | 0.53 | 0.52 | 0.52 |
| AnoGAN | 0.52 | 0.64 | 0.09 | 0.16 | 0.53 |
| GAN-kmeans | 0.5 | 0.5 | 0.45 | 0.48 | 0.5 |
| **Proposed Method** | **0.86** | **0.84** | **0.89** | **0.86** | **0.86** |

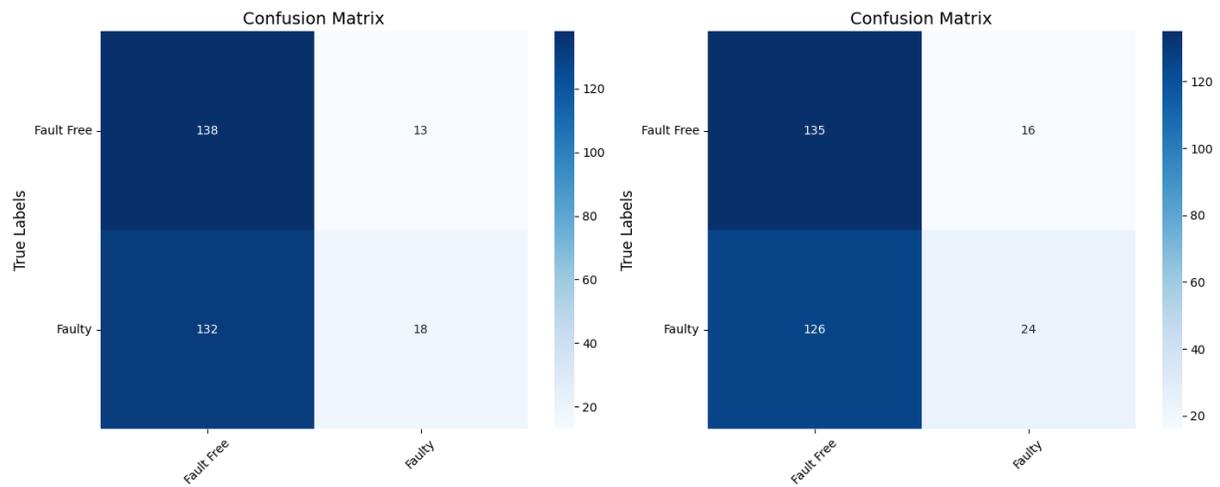
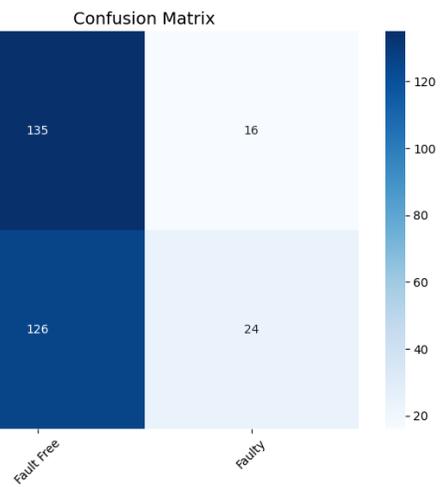

(a) CVAE-T²      (b) CVAE-SPE

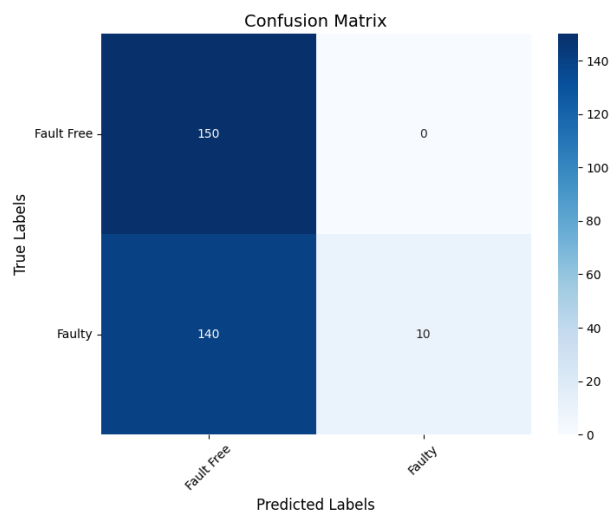
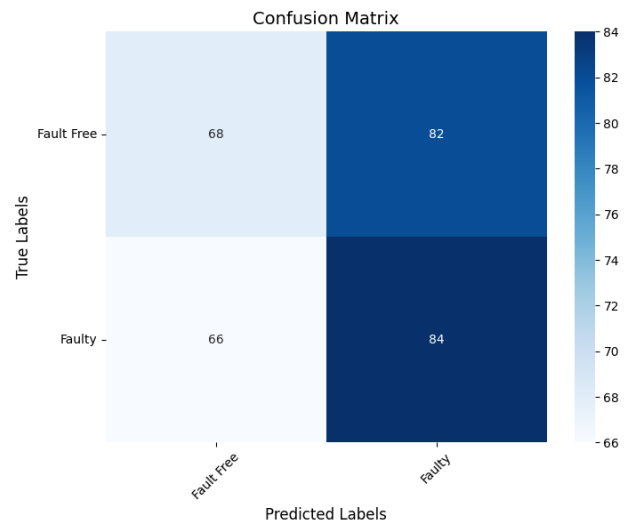

(c) Resnet-DBSCAN      (d) Resnet-Kmeans

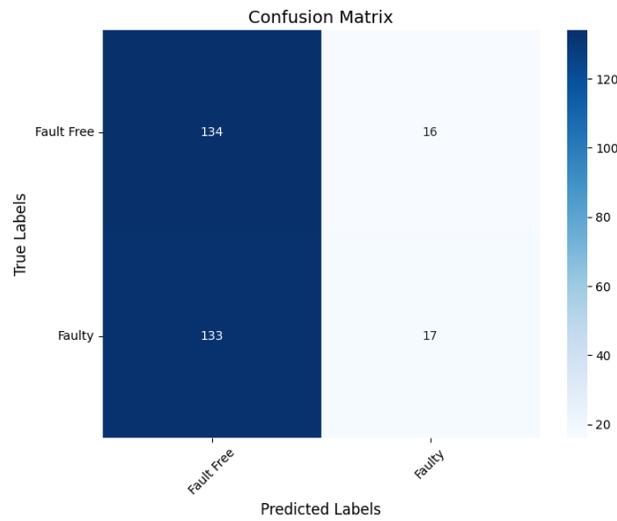
(e) AE-SSIM

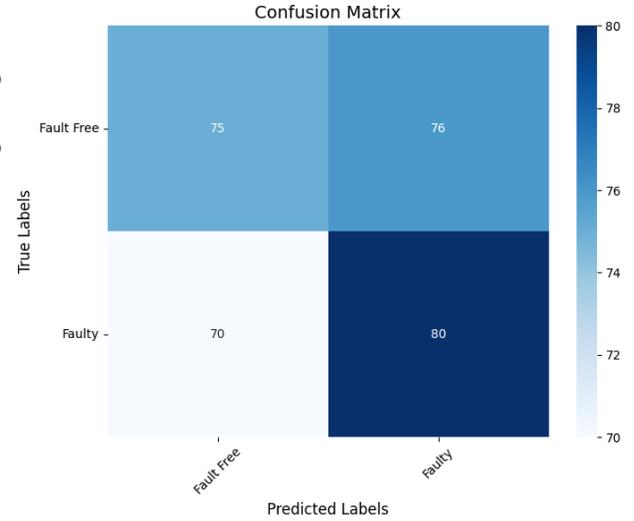
(f) DEC

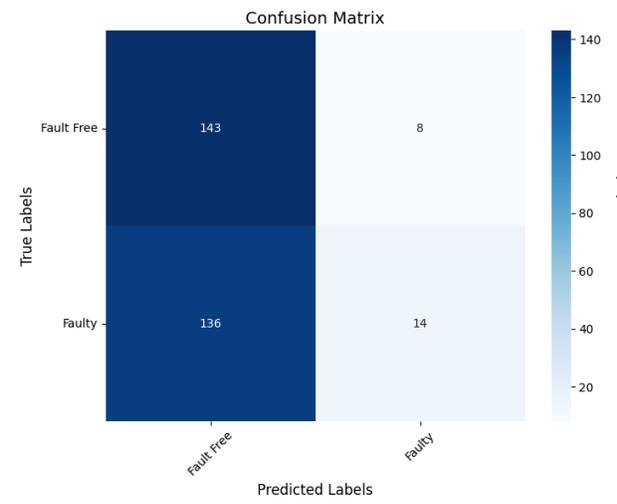
(g) AnoGAN

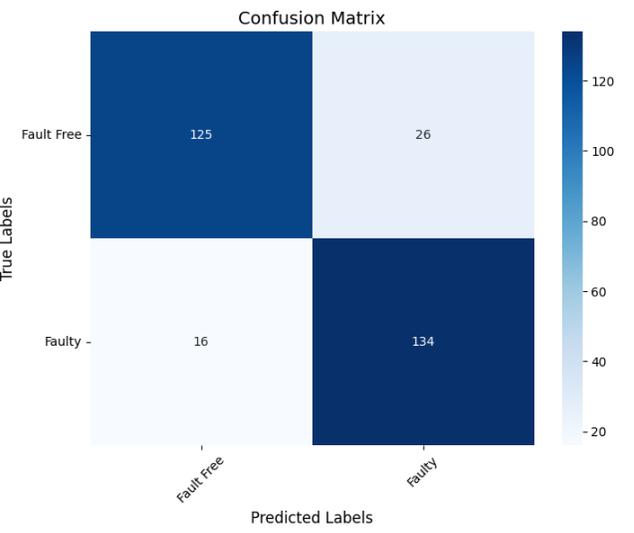
(h) Proposed method

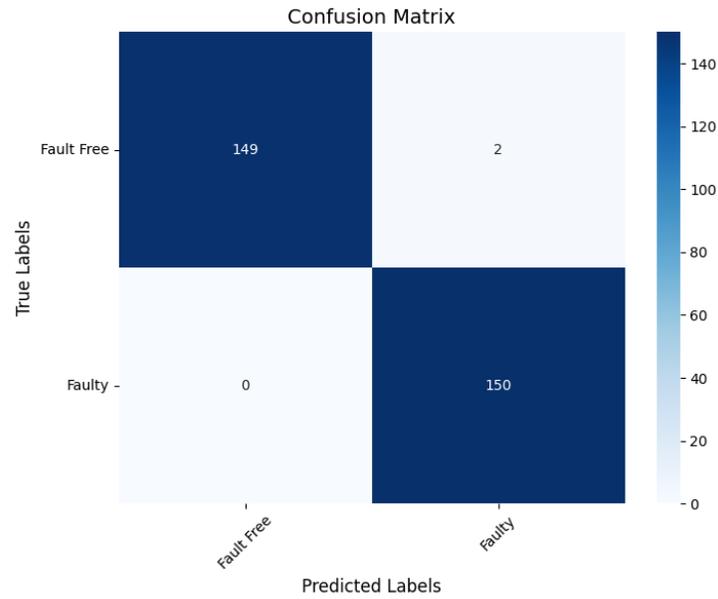

(a) Proposed method + EWMA

Figure 12. Confusion Matrix for all the models applied to dataset I

We also present the results (Table 4 and Figure 13) of the proposed method when combined with the EWMA scheme under the assumption of continuous monitoring. The EWMA approach improves performance, achieving a higher F1-Score of 0.99.

Table 4. Effect of using EWMA on the performance of the proposed model for dataset I

| Method | Accuracy | Precision | Recall | F1-Score | AUROC |
| --- | --- | --- | --- | --- | --- |
| Proposed Method | 0.86 | 0.84 | 0.89 | 0.86 | 0.86 |
| Proposed method-EWMA | **0.99** | **0.99** | **1** | **0.99** | **0.99** |

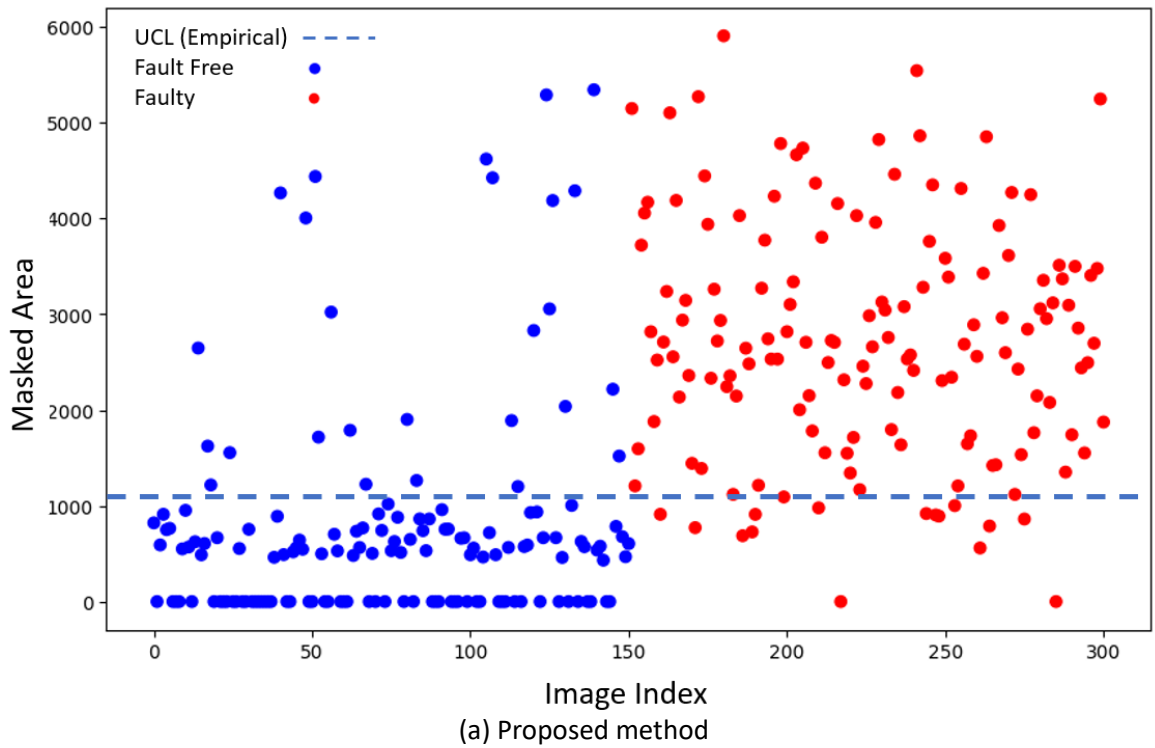

(a) Proposed method

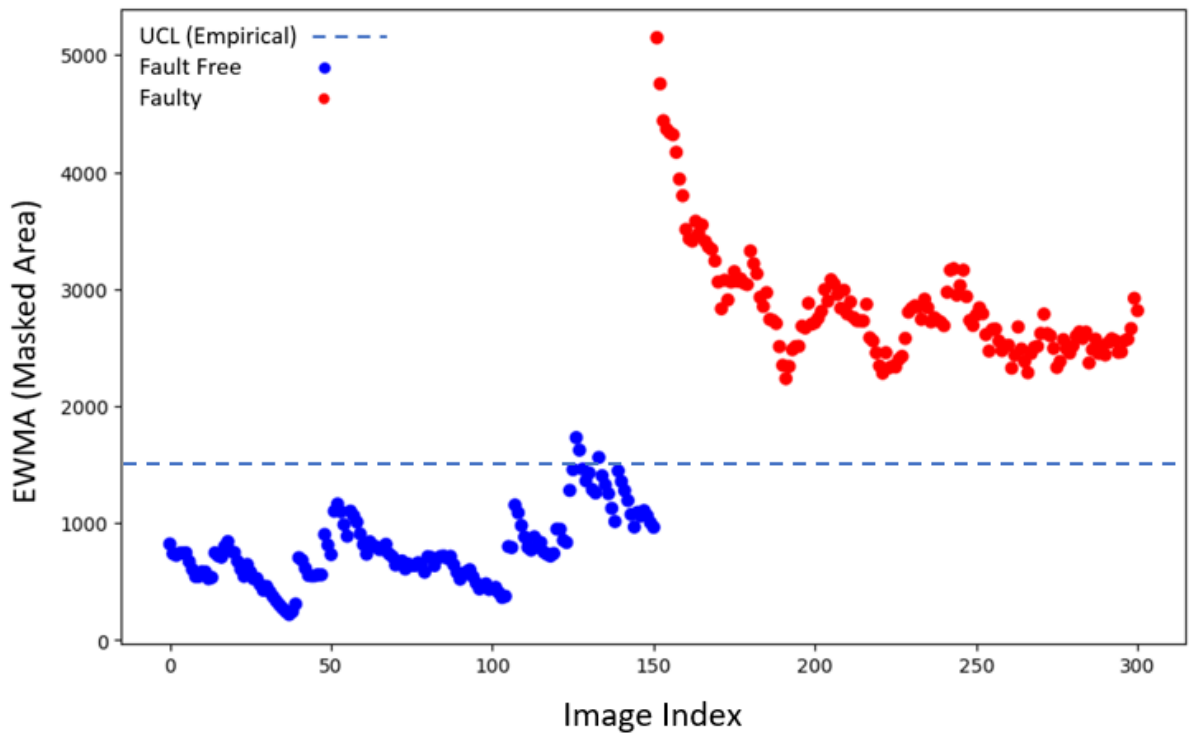

(b) Proposed method + EWMA

Figure 12. Monitoring using the proposed method and EWMA

## 3.2 Open-Source Dataset II

This dataset represents another set of images from the same open-source dataset in the previous section. Sample images from the dataset is seen in Figure 13.

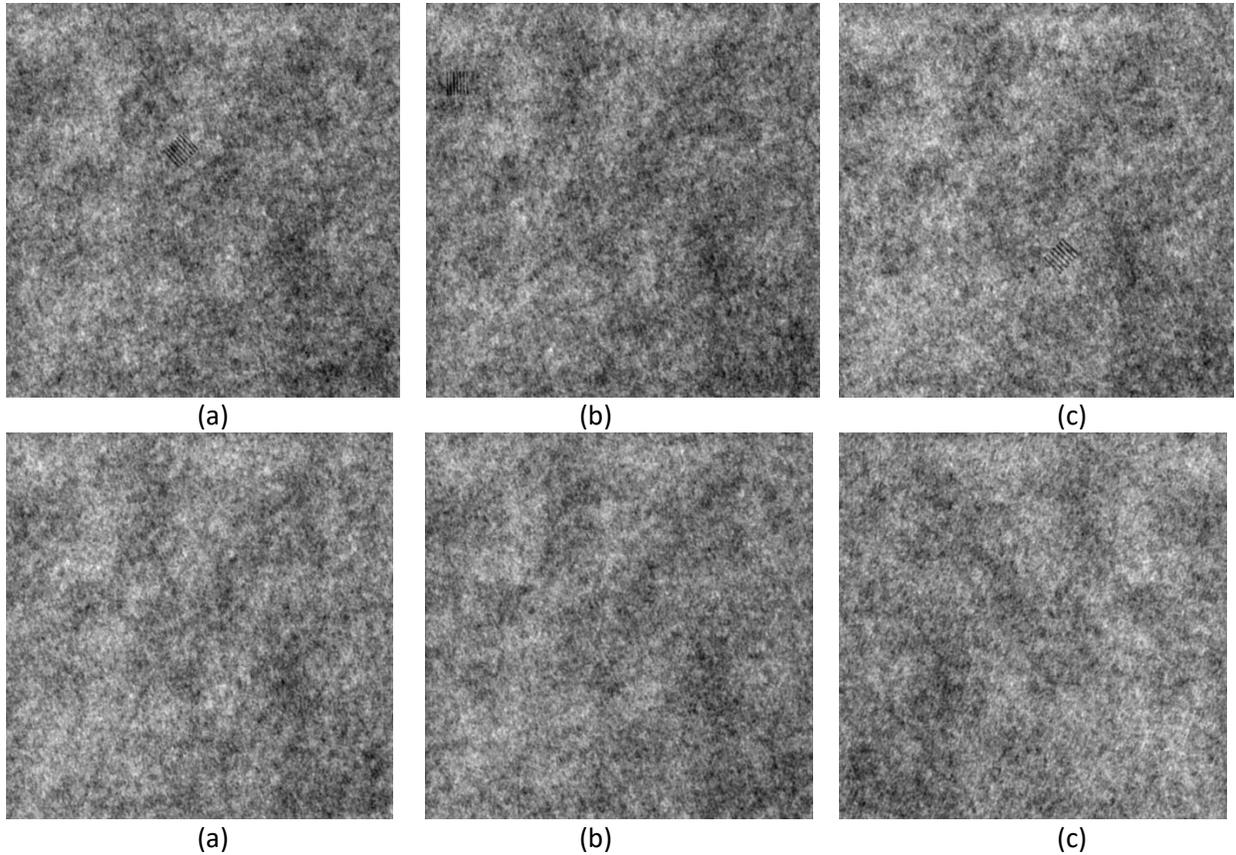

Figure 14. Sample images from dataset II, images a, b, c represent fault free samples, and images d, e, f represent faulty samples

The results from Table 5 show that the proposed method performs significantly better than the others. It achieves the highest accuracy (0.86), precision (0.94), recall (0.78), F1-score (0.85), and AUROC (0.86). Other methods have much lower performance, with many struggling particularly with recall and F1-score.

Table 5. Comparison across all the used models for dataset II

| Method | Accuracy | Precision | Recall | F1 Score | AUROC |
|---|---|---|---|---|---|
| CVAE-$T^2$ Chart | 0.49 | 0.42 | 0.07 | 0.11 | 0.5 |
| CVAE-SPE Chart | 0.49 | 0.43 | 0.08 | 0.13 | 0.55 |
| Resnet-Kmeans | 0.5 | 0.5 | 0.45 | 0.48 | 0.5 |
| Resnet-DBSCAN | 0.5 | 0 | 0 | 0 | 0.5 |
| AE-SSIM | 0.51 | 0.51 | 0.67 | 0.58 | 0.51 |
| DEC | 0.54 | 0.54 | 0.52 | 0.53 | 0.54 |
| AnoGAN | 0.47 | 0 | 0 | 0 | 0.38 |
| GAN-kmeans | 0.51 | 0.51 | 0.51 | 0.50 | 0.51 |
| Proposed Method | **0.86** | **0.94** | **0.78** | **0.85** | **0.86** |

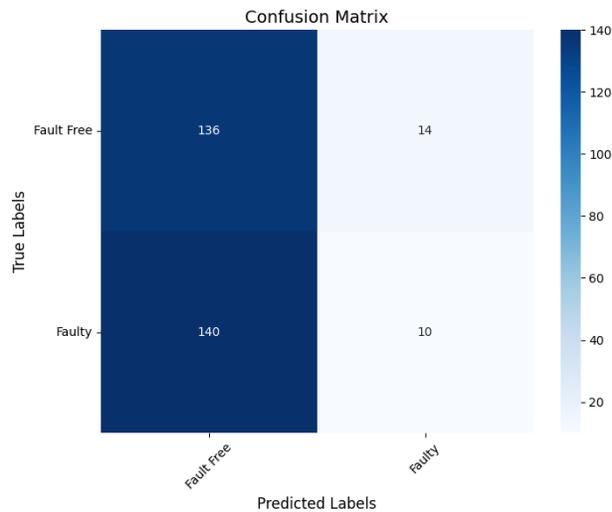

(a) CVAE-T²

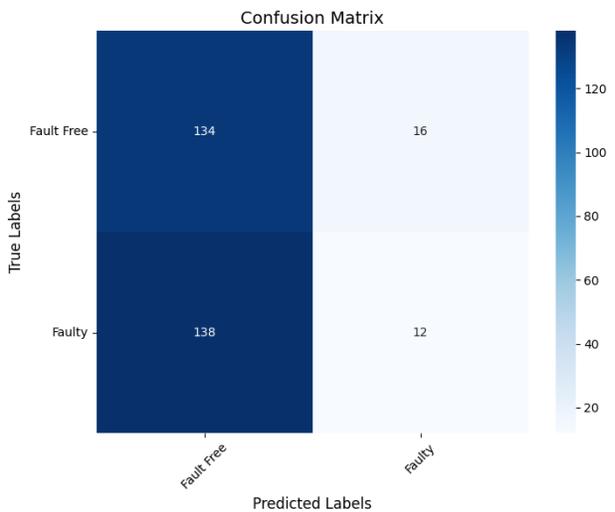

(b) CVAE-SPE

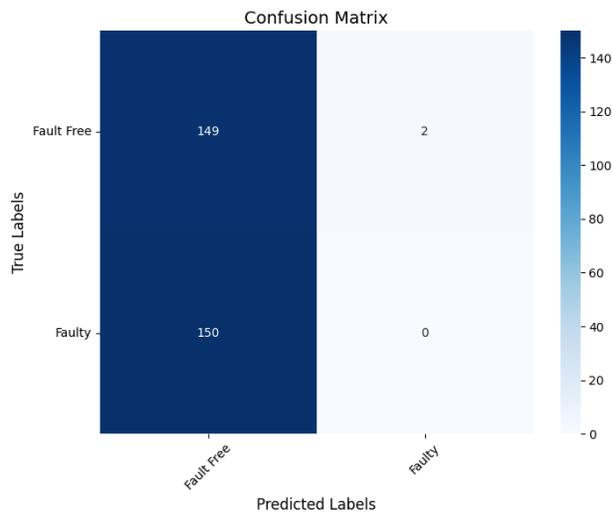

(c) Resnet-DBSCAN

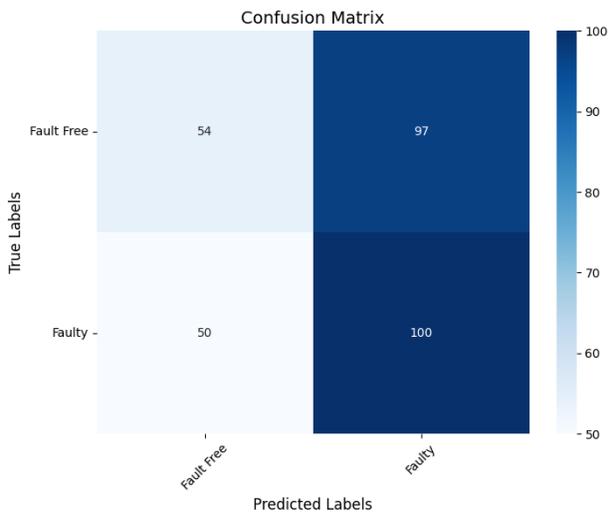

(d) Resnet-Kmeans

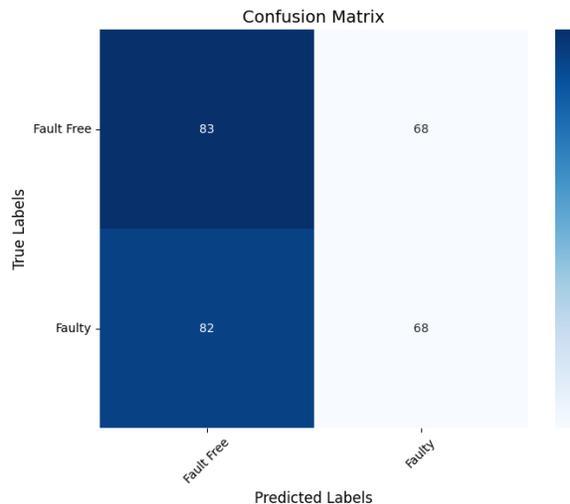
(e) AE-SSIM

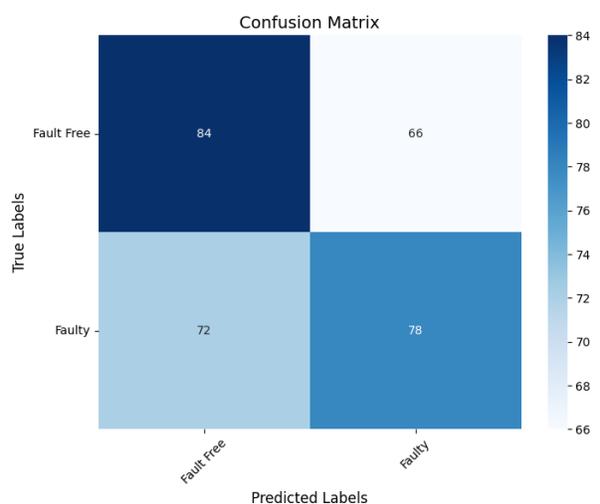
(f) DEC

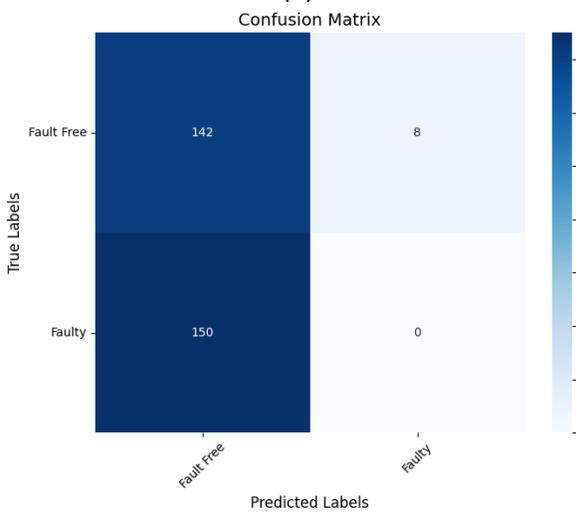
(g) AnoGAN

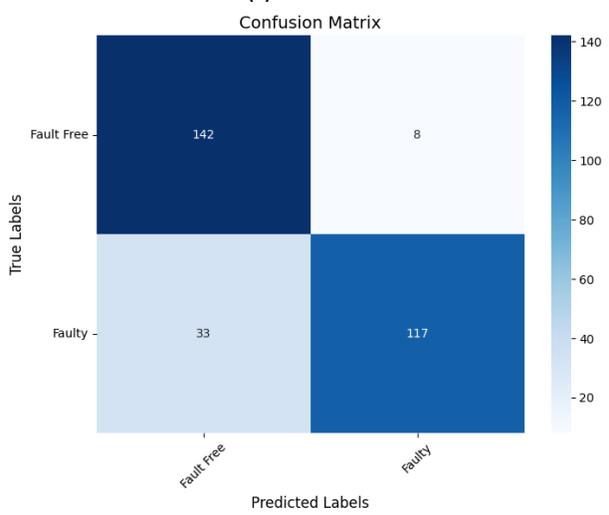
(h) Proposed method

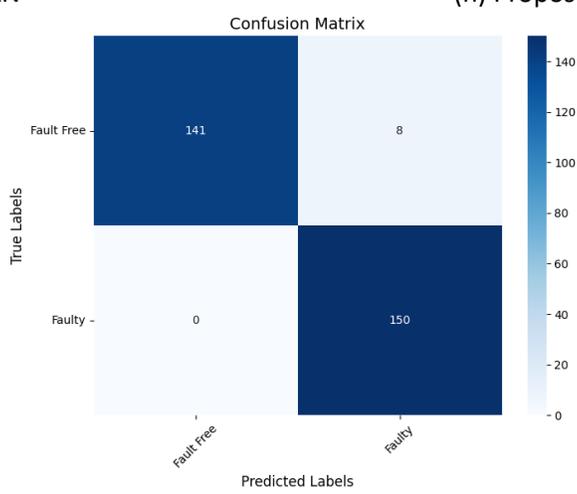
(it) Proposed method + EWMA

Figure 14. Confusion Matrix for all the models applied to dataset II

Table 6. Effect of using EWMA on the performance of the proposed model for dataset II

| Method | Accuracy | Precision | Recall | F1 Score | AUROC |
|---|---|---|---|---|---|
| Proposed Method | 0.86 | 0.94 | 0.78 | 0.85 | 0.86 |
| **Proposed method-EWMA** | **0.97** | **0.94** | **1** | **0.97** | **0.97** |

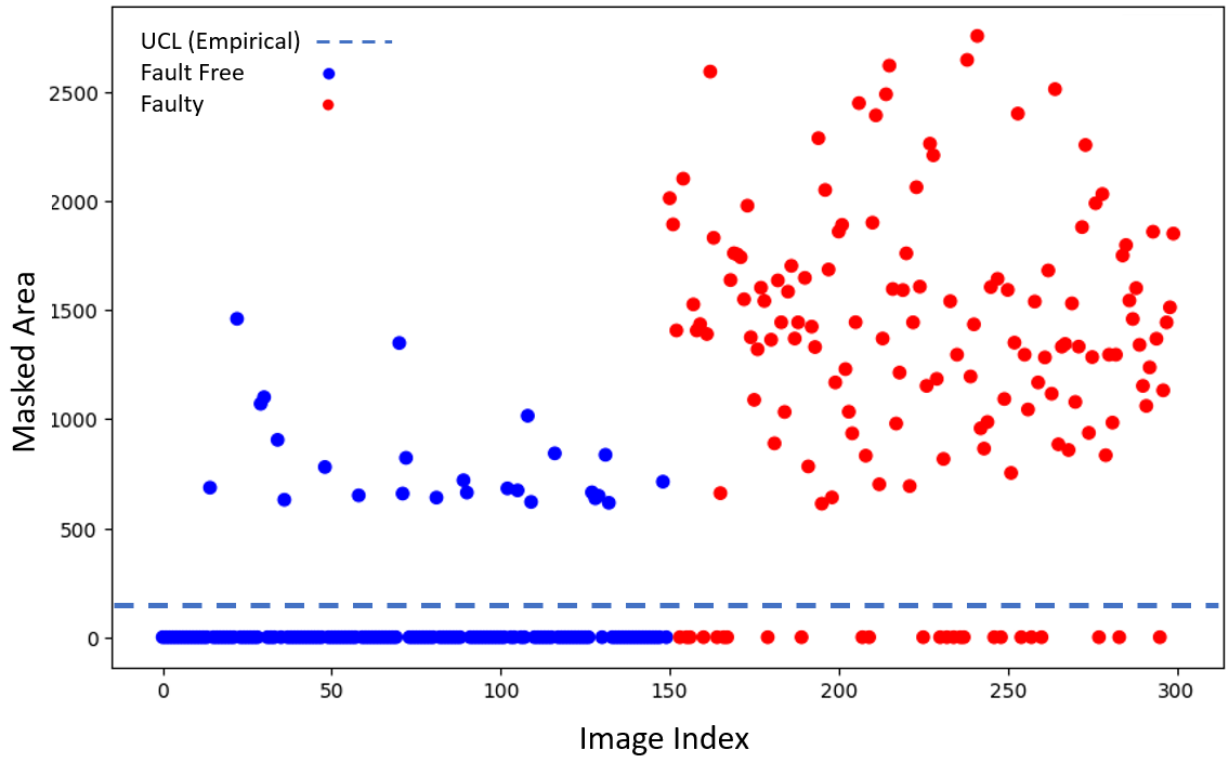

(a) Proposed method

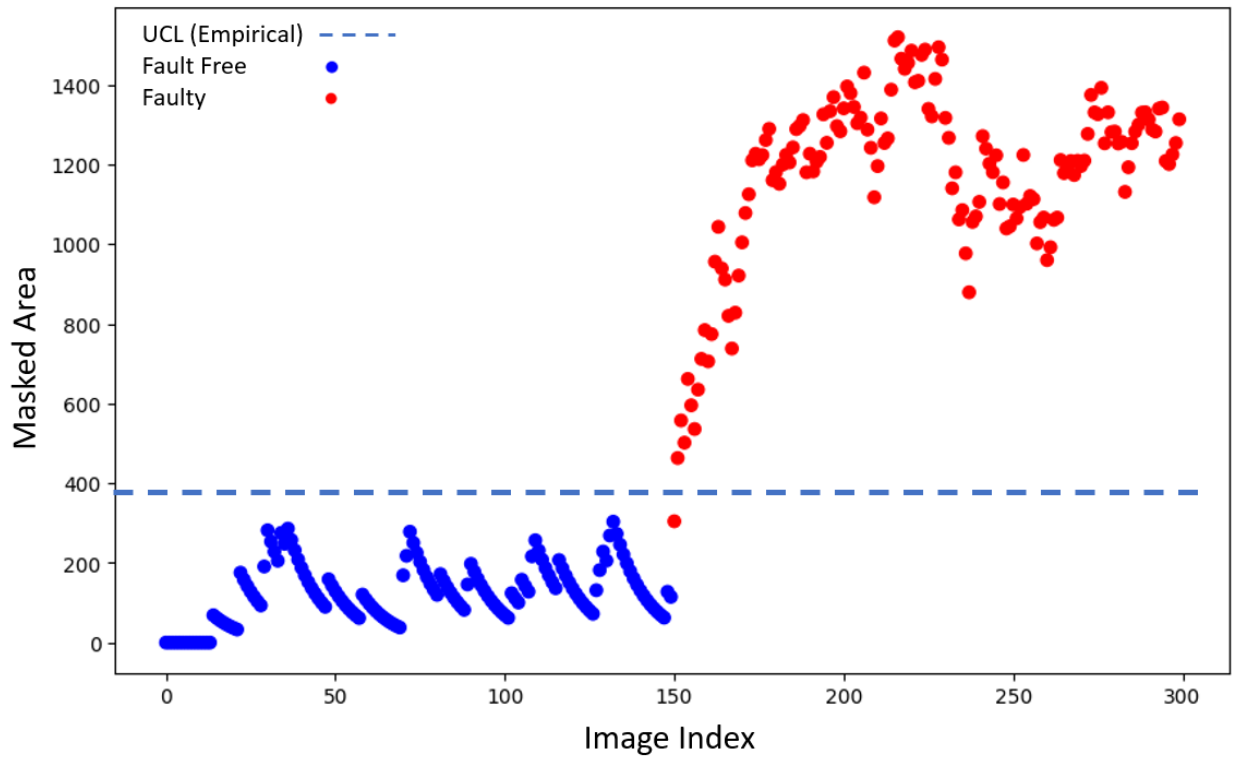

(b) Proposed method + EWMA
Figure 15. Monitoring using the proposed method and EWMA

## 4. Discussions and Recommendations for Parameter Selection

Our proposed method for unsupervised fault detection and monitoring utilizes SAM with a moving window approach to improve segmentation accuracy and reliability. By processing smaller sub-images, SAM can better focus on localized details, enhancing its ability to identify defects. This method is particularly useful in industrial settings where labeled data is scarce or unavailable. The adaptive clustering algorithm further refines the detection process by filtering out noise and identifying consistent defect regions across multiple windows.

When visualizing the latent space using the CVAE with Principal Component Analysis (PCA) and t-distributed Stochastic Neighbor Embedding (t-SNE), we observe significant overlap and a lack of clear clustering in the plots generated by existing methods. This reveals the difficulty in differentiating between defect and non-defect regions. This reinforces the need for more robust and effective strategies, such as our proposed unsupervised fault detection and monitoring approach. In all the datasets used, our method showed excellent ability to distinguish between the two classes unsupervisedly.

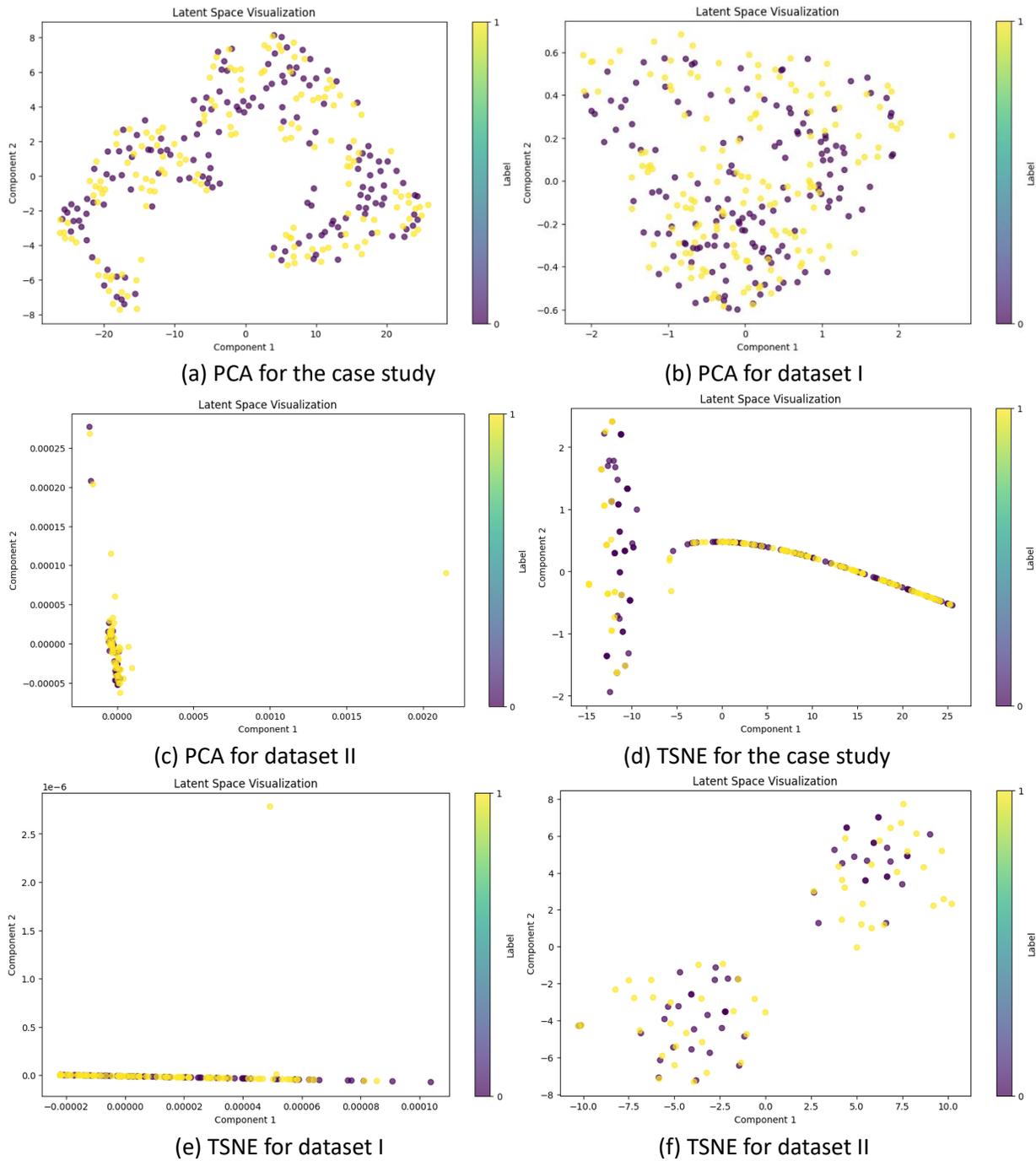

Figure 16. Plot of the latent space using PCA and TSNE for all the used data

Our method is highly explainable and can detect defects accurately. This indicates that users can easily understand and interpret the process and results. The segmentation and clustering steps are transparent, allowing operators to see how each decision is made. This clarity ensures that the model's outputs are not just black box results but are supported by clear, logical steps that demonstrate how defects are identified and categorized. Such explainability is crucial in industrial settings where understanding the reasoning behind defect detection can help troubleshoot and improve the overall quality control process.

Note that the proposed method works under two main assumptions:

- It assumes uniform defect characteristics, meaning the defects to be detected are relatively uniform in size and appearance within each image. This assumption aids in setting appropriate window sizes and thresholds for segmentation and clustering.
- The method assumes defects are localized and not spread across large areas. The moving window approach focuses on smaller sub-images, meaning large, dispersed defects may need to be captured more effectively. These assumptions are critical for the effectiveness of our segmentation and clustering processes.

Most importantly, we have discussed some considerations when selecting the model hyperparameters.

**4.1 Selecting the Window Size and Step Size for Unsupervised Fault Detection and Monitoring**

Selecting the appropriate window size and step size is a multifaceted decision that significantly impacts the effectiveness of our unsupervised fault detection and monitoring method. By considering the nature of defects, image resolution, computational resources, and SAM's performance characteristics, we can optimize these parameters for enhanced accuracy and efficiency. The window size determines the size of the sub-images that SAM processes. Smaller windows allow SAM to focus on localized details, potentially leading to more accurate segmentation of defects. However, if the window size is relatively large, SAM may miss smaller defects or produce less precise segmentation masks. Conversely, suppose the window size is too small. In that case, the method might become computationally expensive due to the increased number of windows to process, and very small windows might not capture the context needed for accurate defect identification.

On the other hand, the step size controls the overlap between consecutive windows. A smaller step size results in more overlap, providing a finer granularity for the analysis, which can enhance defect detection by ensuring that defects located at the edges of windows are not missed. Nonetheless, a smaller step size also increases the number of windows to process, leading to higher computational costs. A larger step size reduces computational load but increases the risk of missing defects that fall between the non-overlapping regions of the windows.

In addition, the selection of window and step sizes is influenced by several critical factors. The nature of defects plays a pivotal role. If the defects are large and well-defined, larger windows and step sizes might suffice. Conversely, smaller windows and step sizes are necessary for smaller, dispersed defects to ensure comprehensive coverage. This adjustment is crucial to capture and segment the defects accurately, thus enhancing detection accuracy. Moreover, image resolution and quality also significantly impact the choice of window size. High-resolution images permit the use of smaller windows without losing detail, but this advantage comes at the cost of increased processing power. On the other hand, lower-resolution images might require larger windows to capture sufficient context for effective segmentation. Therefore, balancing the window size to match the image resolution while maintaining processing efficiency is essential.

The available computational resources, such as processing power and memory, impose practical constraints on the window and step size selection. It is essential to balance the need for detailed analysis with the limitations of the processing environment to optimize performance. Insufficient resources can lead to slower processing times and reduced efficiency, making it crucial to adjust the window and step sizes accordingly.

Understanding the performance characteristics of SAM on different window sizes can greatly inform the selection process. Empirical testing is highly advisable. By conducting experiments with various window and step sizes on a representative sample of surface images, one can gain valuable insights. Analyzing segmentation results, processing time, and accuracy for different configurations helps in identifying the optimal parameters that maximize segmentation accuracy without incurring excessive computational costs. This iterative testing and validation process is key to fine-tuning the method for robust and reliable defect detection in diverse industrial settings.

### 4.2 Tolerance and Thresholds in the Clustering Algorithm

Tolerance and thresholds are important for fine-tuning the results from the model. The thresholds are used to filter out irrelevant segmentation masks based on their area. Masks with areas falling within the defined range are considered potentially significant and are included in further analysis. The lower threshold ensures that very small areas, likely representing noise or insignificant features, are excluded from consideration. The upper threshold helps eliminate very large areas that may not correspond to individual defects or may represent larger non-defective regions.

Tolerance is used in the clustering algorithm to determine whether areas are close enough to be considered part of the same cluster. It defines the acceptable deviation from the mean area of the current cluster. A smaller tolerance results in tighter clusters, ensuring that only very similar areas are grouped together. This can improve the precision of defect identification but may exclude some relevant areas. A larger tolerance allows for broader clusters, capturing more areas but increasing the risk of including dissimilar areas, which can reduce the specificity of defect detection.

For defects that are relatively uniform in size, narrower thresholds and smaller tolerance may be appropriate. For defects with greater size variability, broader thresholds and larger tolerance may be necessary to capture all relevant areas. Tolerance, on the other hand, is highly dependent on window size and step size.

### 4.3 EWMA Scheme and Smoothing Parameter Selection

The integration of the Exponentially Weighted Moving Average (EWMA) scheme adds an additional layer of robustness to our method. EWMA allows for continuous monitoring and analysis of defect trends, ensuring that even small changes in defect patterns are detected. This continuous feedback loop is critical for proactive maintenance and quality control in industrial applications. Our results demonstrate that this combined approach can achieve high detection accuracy and reliability, making it a valuable tool for various industrial monitoring tasks.

The smoothing parameter in the EWMA scheme is of utmost importance, as it determines the weight given to recent observations versus past data and optimizes its performance. A smaller smoothing parameter places more emphasis on recent data, making the method more responsive to new defects. However, this can also lead to increased sensitivity to noise. On the other hand, a larger smoothing parameter gives more weight to historical data, which can smooth out short-term fluctuations but may delay the detection of new defects. The user can choose any values that fall between 0 and 1 based on the application, with typical values usually used being 0.1 and 0.2.

### 5. Conclusions, Limitations and Future Directions

This paper presented a novel method for unsupervised fault detection and monitoring, utilizing the Segment Anything Model (SAM) with a moving window approach. This method addresses the significant challenge of detecting surface defects without the need for labeled data, making it particularly useful in industrial applications where such data is often scarce or difficult to obtain. Our approach significantly improves segmentation accuracy by dividing images into smaller sub-images and processing each individually through SAM. This is further refined by calculating the areas of segmented regions and employing an adaptive clustering algorithm to identify consistent defect regions while filtering out noise.

The integration of the Exponentially Weighted Moving Average (EWMA) scheme enhances the method's capability by enabling continuous monitoring and analysis of defect trends over time. This continuous feedback loop is critical for predictive/proactive maintenance and quality control in industrial applications. Our results demonstrate that this combined approach can achieve high detection accuracy and reliability, making it a valuable tool for various industrial monitoring tasks and has potential to significantly reduce manual labor and improve the production optimization, relieving the burden on industrial professionals.

On the limitations side, the moving window approach, while improving accuracy, increases computational load due to the need to process multiple sub-images. This can be resource-intensive, especially for high-resolution images or extensive datasets. Additionally, the choice of window size and step size is crucial for the method's effectiveness. Inappropriate settings can lead to either missed defects or unnecessary computational overhead. Therefore, empirical testing and validation on a representative sample of surface images are essential to optimize these parameters.

Another limitation is the sensitivity of the predefined upper and lower thresholds for mask area calculation. If these thresholds are not well-calibrated, the method may either include too much noise or miss significant defective areas. SAM, although effective for segmentation, sometimes generates unwanted masks, which can affect overall detection accuracy. Using prompts could mitigate this issue, but more experiments are required to validate this.

Another aspect is that processing smaller windows can sometimes capture irrelevant details or artifacts, which may be mistaken for defects. While the clustering algorithm helps reduce this noise, it does not eliminate it entirely. Implementing this method in a real-time industrial setting may pose challenges due to the need for rapid processing and analysis. Ensuring the method can operate within the required time constraints without compromising accuracy is essential.

Future work should focus on optimizing the algorithm's efficiency, possibly through parallel processing or more efficient coding practices. Additionally, further testing and validation on a wider range of industrial datasets will help generalize the method's applicability and reliability. Implementing adaptive strategies where thresholds and tolerance are dynamically adjusted based on the characteristics of the input images can improve detection accuracy.

Moreover, employing a multi-stage filtering approach is a very interesting idea. It can also enhance defect detection. Broad initial thresholds can capture a wide range of potential defects, followed by more stringent thresholds to refine the analysis in subsequent stages. Another exciting direction is context-aware adjustments, considering the defects' location on the surface, which can inform adjustments to thresholds and tolerance, with defects in critical areas warranting stricter settings. Machine learning techniques (e.g., k-means and DBSCAN) to predict optimal thresholds and tolerance based on historical data can automate the selection process. Finally, for cluster selection, considering the minimum number of points in the cluster, rather than just relying on the maximum cluster, can improve defect detection accuracy.